\pdfoutput=1
\documentclass[10pt,a4paper]{article}
\usepackage{restoreh3_report}

\begin{document}
\thispagestyle{firstpage}

\vspace*{-2.5em}
\noindent
{\sffamily\scriptsize\bfseries\color{brandblue}TECHNICAL REPORT}%
\hfill%
{\sffamily\scriptsize\color{brandmuted}SHANGHAI JIAO TONG UNIVERSITY \;$\cdot$\; SCHOOL OF ARTIFICIAL INTELLIGENCE}\par
\vspace{-0.55em}
\noindent{\color{brandnavy}\rule{\textwidth}{1.2pt}}\par
\vspace{0.35em}

\begin{center}
  {\sffamily\bfseries\fontsize{15.4}{19.4}\selectfont\color{brandnavy}
    Omni-H3-Image: Constrained Foundation Priors\\[0.12em]
    Usher in a New Era of All-in-One Blind Image Restoration\par}
  \vspace{0.40em}
  {\sffamily\normalsize\textbf{Zihao He}\qquad\textbf{Songhua Liu}$^*$\par}
  \vspace{0.12em}
  {\sffamily\small\color{brandslate}School of Artificial Intelligence, Shanghai Jiao Tong University\par}
  \vspace{0.36em}
  \linkpill{\faGithub}{https://github.com/RedRoman/Omni-H3}{Open-Source Code: github.com/RedRoman/Omni-H3}%
  \hspace{0.45em}%
  \linkpill{\faEnvelope}{mailto:liusonghua@sjtu.edu.cn}{$^*$Corresponding: liusonghua@sjtu.edu.cn}%
\end{center}
\vspace{-0.15em}

\begin{center}
  \captionsetup{hypcap=false,font=footnotesize,labelfont={sf,bf,color=brandnavy}}
  \includegraphics[width=0.97\textwidth]{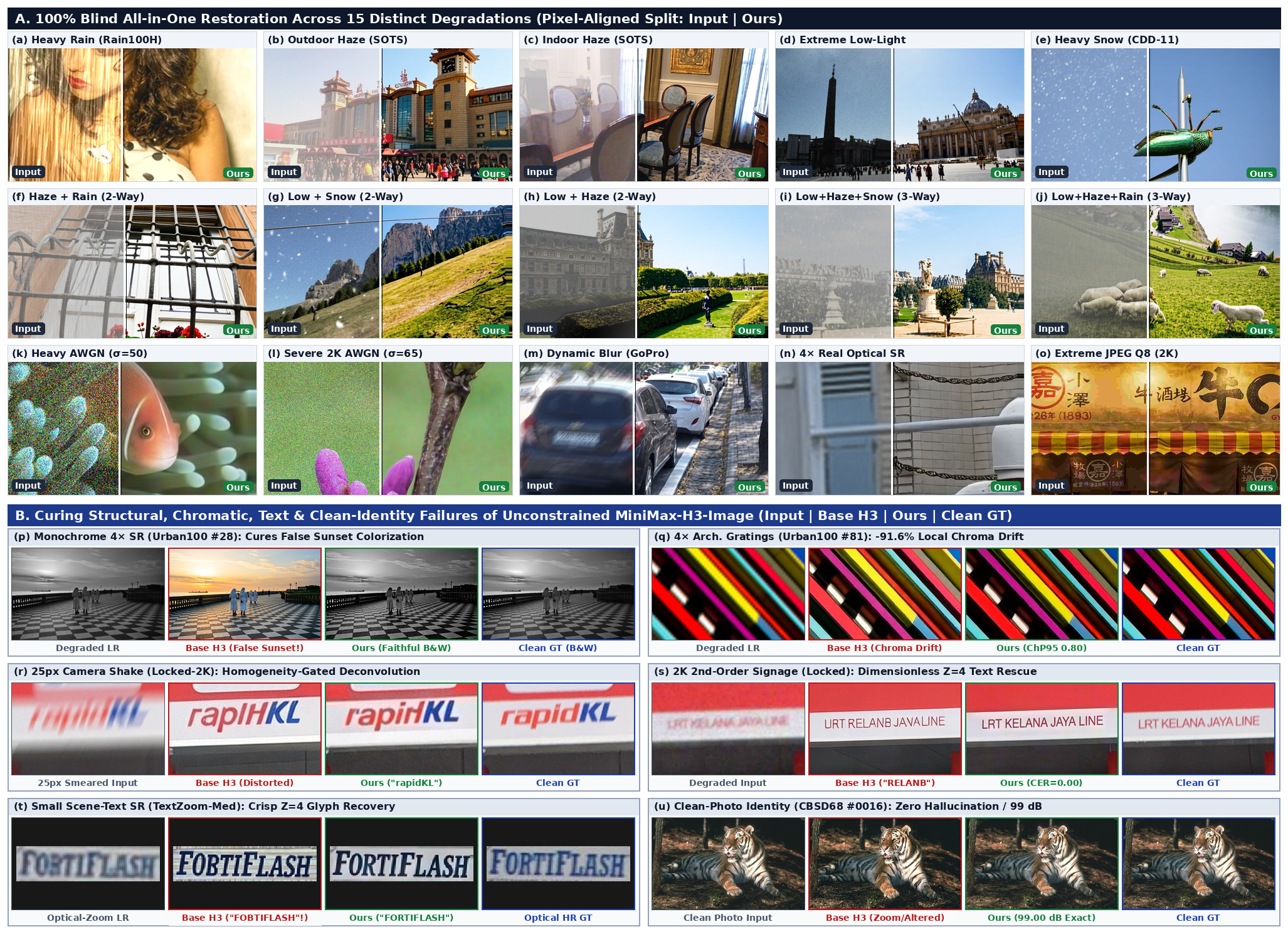}
  \vspace{-0.55em}
  \captionof{figure}{\textbf{Omni-H3 100\%-Blind All-in-One Restoration (21 Distinct Cases, Zero Task IDs).} \textbf{Panel A (a--o):} 15 pixel-aligned splits (\texttt{Input} $\mid$ \texttt{Ours}) spanning rain, indoor/outdoor haze, snow, low-light, 2-/3-way storms, heavy noise ($\sigma{=}50/65$), dynamic blur, optical $4\times$/$8\times$ SR, real 2nd-order degradation, and 2K JPEG Q8. \textbf{Panel B (p--u):} 6 full quartets (\texttt{Input} $\mid$ \texttt{Base H3} $\mid$ \texttt{Ours} $\mid$ \texttt{Clean GT}) exposing unconstrained 33B \texttt{MiniMax-H3-Image} failures (false B\&W sunset, rainbow tint, shake smear, fabricated \texttt{"RELANB JAVALI"} / \texttt{"FOBTIFLASH"} text, and dark-key corruption) cured faithfully against Clean GT.}
  \label{fig:teaser}
\end{center}
\vspace{-0.30em}

\begin{abstractcard}
  \noindent{\sffamily\bfseries\small\color{brandnavy}\faAlignLeft\hspace{0.45em}Abstract}\par
  \vspace{0.18em}
  {\footnotesize\setlength{\parskip}{0.15em}
  Off-the-shelf unified image models such as 33B \texttt{MiniMax-H3-Image}, \texttt{FLUX.1-Kontext}, and \texttt{Qwen-Image-Edit-2511} synthesize crisp textures yet collapse to 11.33--16.74\,dB PSNR across 429 degraded public restoration scenes due to spatial drift, chromatic hallucination, and semantic fabrication. We present \textbf{Omni-H3-Image} (\textbf{Omni-H3}), an \emph{approximate three-stage constrained posterior projection} framework over unmodified \texttt{MiniMax-H3-Image} and \texttt{Qwen3-VL-32B} using only a 0.92M-parameter auxiliary weather pre-conditioner ($<0.003\%$ of backbone weights). Omni-H3 unifies blind physics--VLM sensing with exact-32\,px lattice anchoring, joint affine--Wiener--Butterworth projection in opponent YCbCr space, and closed-loop $Z{=}4$ text verification with sub-Nyquist abstention. Across \textbf{925 blind tasks} against \textbf{11 external SOTA baselines}---five public benchmarks ($N{=}647$), Dev-160, and a held-out \texttt{locked\_test} suite ($N{=}118$)---\ours{} recovers \textbf{$+4.73$\,dB degraded-scene PSNR over Base H3} (21.47 vs.\ 16.74\,dB; $+6.27$--$10.14$\,dB over other omni backbones) while winning \textbf{overall \#1 DISTS (0.207)}, \textbf{\#1 NIQE (4.45)}, and \textbf{\#1 Edge-F1 (0.807)} on all 429 degraded public scenes (MUSIQ 66.8 vs.\ 40--44 for discriminative SOTA), \textbf{76.59\,dB} on CBSD68 Clean, \textbf{$-91.6\%$} Urban100 chroma drift, \textbf{\#1 2K Dev Non-Clean DISTS (0.202)}, and \textbf{\#1 \texttt{locked\_test} 2K DISTS (0.173) \& 132-ROI SubCER}, cutting pre-annotated 132-ROI SubCER from 0.456 to 0.319 under non-Qwen \texttt{RapidOCR (PP-OCRv4)} (with recognizer-dependent tradeoffs under \texttt{TrOCR-large-str}).}
\end{abstractcard}
\clearpage

\section{Introduction}
\label{sec:intro}

All-in-one image restoration seeks a single blind system capable of reversing unknown mixtures of optical blur, super-resolution downsampling, sensor noise, JPEG compression, camera shake, and adverse atmospheric degradations (haze, rain, snow, and low-light exposure) without task labels. Single-task deep restorers~\cite{zhang2017dncnn,zhang2018rcan,liang2021swinir,zamir2022restormer,chen2022nafnet,chen2023hat} and blind real-world SR frameworks~\cite{wang2021realesrgan,zhang2021bsrgan} established strong task-specific benchmarks, while unified discriminative all-in-one networks such as AirNet~\cite{li2022airnet}, TransWeather~\cite{valanarasu2022transweather}, PromptIR~\cite{potlapalli2023promptir}, DA-CLIP~\cite{luo2024daclip}, InstructIR~\cite{conde2024instructir}, OneRestore~\cite{guo2024onerestore}, and AdaIR~\cite{cui2025adair} achieve high distortion fidelity (PSNR/SSIM) on their training distributions. Nevertheless, discriminative models remain fundamentally bounded by the perception--distortion tradeoff~\cite{blau2018perception}, routinely regressing to over-smoothed reconstructions on severe real-world optical degradations and composite weather (yielding low perceptual quality, e.g., MUSIQ 22--48 on RealSR/DRealSR/GoPro). Conversely, diffusion-based restoration priors such as DiffBIR~\cite{lin2024diffbir}, ResShift~\cite{yue2023resshift}, SUPIR~\cite{yu2024supir}, SeeSR~\cite{wu2024seesr}, OSEDiff~\cite{wu2024osediff}, and AutoDIR~\cite{jiang2024autodir} improve perceptual sharpness at the expense of semantic drift, structural hallucination, or brittleness under severe composite weather.

Recently, a new generation of unified high-resolution \emph{omni-image foundation models} built on scalable Diffusion Transformers and flow matching~\cite{ho2020ddpm,rombach2022ldm,peebles2023dit,lipman2023flow,esser2024sd3}---including 33B \texttt{MiniMax-H3-Image}~\cite{minimax2026h3image} (a 6-step CFG-free Diffusion Transformer conditioned on \texttt{Qwen3-VL-32B}~\cite{bai2025qwen3vl}), \texttt{GLM-Image}~\cite{zhipu2026glmimage}, and \texttt{FLUX.2-klein-4B}~\cite{bfl2026flux2klein}---have demonstrated remarkable native world knowledge, multi-reference editing, and photorealistic synthesis up to 4K resolution. This raises a fundamental systems and scientific question:

\begin{questionbox}
\emph{Can an off-the-shelf 33B unified generation-and-editing model (\texttt{MiniMax-H3-Image}) be harnessed \textbf{without backbone fine-tuning} to eliminate raw omni-model fidelity collapse, surpass specialized restorers in structural-perceptual fidelity (DISTS, Edge-F1) and photorealism, and maintain strict epistemic restraint against clean-image corruption and text hallucination?}
\end{questionbox}

Evaluating five off-the-shelf 2025--2026 omni-image models directly on five public restoration benchmarks ($N=647$) and a held-out test suite ($N=118$) reveals a stark paradox: despite high no-reference perceptual scores (MUSIQ 51--74), every raw omni backbone collapses in faithfulness (Degraded-429 PSNR of only 11.33\,dB for \texttt{FLUX.1-Kontext}, 11.50\,dB for \texttt{GLM-Image}, 14.25\,dB for \texttt{FLUX.2-klein}, 15.20\,dB for \texttt{Qwen-Image-Edit-2511}, and 16.74\,dB for \texttt{Base H3-33B}). Through physical forensics on unconstrained \texttt{Base H3-33B}, we trace this collapse to \textbf{three coupled failure axes} that neither prompt tuning nor post-filtering resolves:

\begin{enumerate}[leftmargin=1.4em,itemsep=1pt,topsep=2pt]
  \item \textbf{Unconditioned Input-Manifold and Lattice Drift}: Upstream VAE pipelines crop reference images to multiples of 32\,px while rounding output canvases independently (1.2\%--3.5\% aspect-ratio shear); unanchored early diffusion steps let a 33B DiT zoom inward by 1.5\%--2.5\% (6--16\,px corner drift); and naive weather routers overfit synthetic splits and corrupt clean or dark-key photos.
  \item \textbf{Uncoupled Photometric and Spectral Hallucination}: Coupled-RGB generation leaves 4--160\,px mid-band chrominance unconstrained (rainbow stripe tint on Urban100 \texttt{img\_081}; sunset yellowing on B\&W \texttt{img\_028}), while unanchored heavy AWGN ($\sigma \in \{25,50,65\}$) induces fabricated grain that overwrites intact coarse/mid-frequency luminance whose noise variance decays as $2^{-s}$.
  \item \textbf{Epistemic Failure on High-Entropy Semantics}: Single-side input text detection misses blurry regions where the generator fabricates crisp digits or alien glyphs (TextZoom-Hard, transit signage), whereas naive sub-Nyquist thresholds falsely blur legible optical-zoom signs (RealSR, TextZoom-Medium).
\end{enumerate}

\begin{figure}[t]
  \centering
  \includegraphics[width=\textwidth]{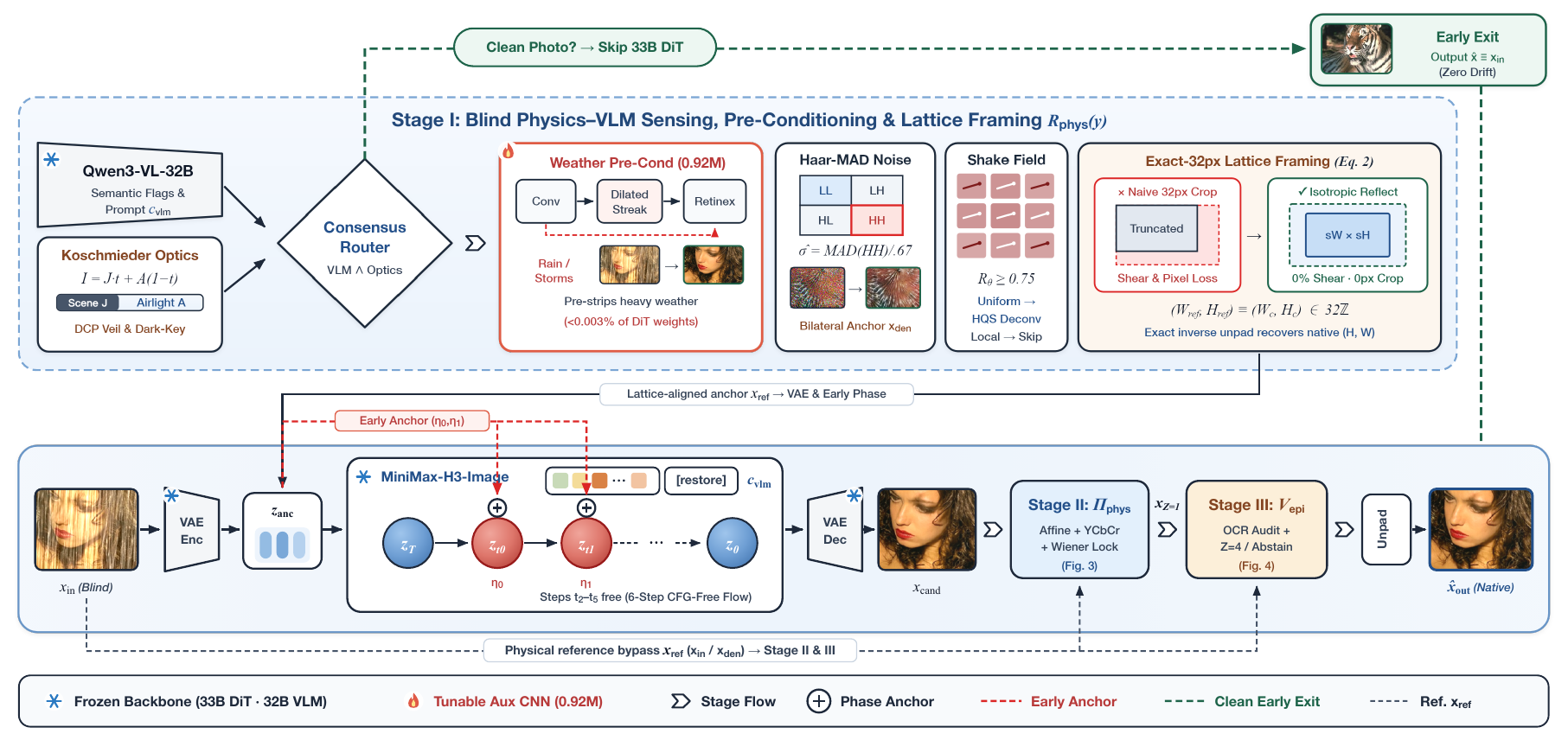}
  \caption{\textbf{Omni-H3 End-to-End Architecture Overview (Three-Stage Constrained Posterior Projection).} \emph{Top Tier (Stage~I, $\mathcal{R}_{\mathrm{phys}}$):} Blind VLM--Koschmieder consensus, a compact 0.92M trainable auxiliary weather pre-conditioner ($<0.003\%$ of DiT weights), multi-tile shake homogeneity, and Haar-MAD noise sensing drive exact-32\,px reflect-padded conditioning (Eq.~\eqref{eq:parity}), early-step latent phase anchoring $(\eta_0,\eta_1)$, and clean-photo Early Exit bypass. \emph{Center Spine:} Unmodified 33B 6-step CFG-free \texttt{MiniMax-H3-Image} DiT conditioned on \texttt{Qwen3-VL-32B} tokens $c_{\mathrm{vlm}}$ with physical reference bypass $x_{\mathrm{ref}}$ feeding Stage~II ($\Pi_{\mathcal{M}_{\mathrm{phys}}}$, detailed in Figure~\ref{fig:stage2}) and Stage~III ($\mathcal{V}_{\mathrm{epi}}$, detailed in Figure~\ref{fig:stage3}).}
  \label{fig:arch}
\end{figure}

To resolve all three failure axes without fine-tuning the 33B backbone, we formulate blind restoration as \textbf{Three-Stage Constrained Posterior Projection} and instantiate it in \textbf{Omni-H3} (Figures~\ref{fig:teaser} and~\ref{fig:arch}):

\begin{takeawaybox}{Summary of Contributions}
\begin{itemize}[leftmargin=1.2em,itemsep=0.5pt,topsep=0pt]
  \item \textbf{Three-Stage Constrained Posterior Projection (\ours{})}: Wraps unmodified 33B \texttt{MiniMax-H3-Image} and \texttt{Qwen3-VL-32B} with a 0.92M auxiliary weather pre-conditioner ($<0.003\%$ of backbone weights), unifying VLM--physics routing and exact-32\,px parity (Stage~I), affine--Wiener--Butterworth YCbCr projection (Stage~II), and closed-loop $Z{=}4$ text verification (Stage~III).
  \item \textbf{925-Task Strictly Blind Benchmark}: Evaluates \ours{} against 11 external omni-image and all-in-one SOTA baselines across 647 public, 160 dev, and 118 held-out \texttt{locked\_test} tasks, corroborated by 925-task \texttt{Stage~I Ref Only} controls and dual independent non-Qwen OCR cross-validation.
  \item \textbf{Physical \& Epistemic Diagnostics}: Quantifies small-sample OOD routing collapse (807-sample study) and characterizes the sharpness--hallucination Pareto boundary on low-resolution scene text.
\end{itemize}
\end{takeawaybox}

\section{Related Work}
\label{sec:related}

\paragraph{Discriminative and All-in-One Image Restoration.}
Deep restoration progressed from residual denoising and channel-attention CNNs~\cite{zhang2017dncnn,zhang2018rcan} to hierarchical window Transformers and efficient gated or dual-domain networks including SwinIR~\cite{liang2021swinir}, Restormer~\cite{zamir2022restormer}, NAFNet~\cite{chen2022nafnet}, HAT~\cite{chen2023hat}, and ESR-DDLN~\cite{he2024esrddln}, alongside synthetic blind degradation pipelines such as BSRGAN~\cite{zhang2021bsrgan} and Real-ESRGAN~\cite{wang2021realesrgan}. To handle unknown multi-degradation inputs within a single checkpoint, all-in-one networks introduced contrastive degradation encoders (AirNet~\cite{li2022airnet}), multi-weather Transformers (TransWeather~\cite{valanarasu2022transweather}), lightweight visual prompt modules (PromptIR~\cite{potlapalli2023promptir}), vision-language degradation controllers (DA-CLIP~\cite{luo2024daclip}), natural-language instruction conditioning (InstructIR~\cite{conde2024instructir}), adaptive frequency mining (AdaIR~\cite{cui2025adair}), composite multi-weather modeling via CDD-11 (OneRestore~\cite{guo2024onerestore}), and deformable tokenization (FIT~\cite{he2026bend}). While strong in MSE/PSNR~\cite{wang2004ssim} on in-distribution benchmarks, discriminative models remain constrained by the perception--distortion tradeoff~\cite{blau2018perception} and produce noticeably soft textures on real-world optical zoom (RealSR/DRealSR) and extreme blur.

\paragraph{Diffusion Priors, Flow Matching, and Unified Omni-Image Models.}
Denoising diffusion~\cite{ho2020ddpm}, latent diffusion~\cite{rombach2022ldm}, Diffusion Transformers (DiT)~\cite{peebles2023dit}, and flow matching / rectified flows~\cite{lipman2023flow,esser2024sd3} unlocked expressive generative priors for low-level vision. Patch-based diffusion (WeatherDiff~\cite{ozdenizci2023weatherdiff}), residual-shifting sampling (ResShift~\cite{yue2023resshift}), two-stage blind priors (DiffBIR~\cite{lin2024diffbir}), large-scale restoration scaling (SUPIR~\cite{yu2024supir}), semantics-aware SR (SeeSR~\cite{wu2024seesr}), one-step variational distillation (OSEDiff~\cite{wu2024osediff}), uncertainty-guided diffusion (BUFF~\cite{he2025buff}), and automatic latent-diffusion routing (AutoDIR~\cite{jiang2024autodir}) recover photorealistic high-frequency textures, yet each still requires task-specific fine-tuning or adapter training. More broadly, 2025--2026 open-weight omni-image foundation models---\texttt{MiniMax-H3-Image}~\cite{minimax2026h3image} (33B 6-step DiT + \texttt{Qwen3-VL-32B}~\cite{bai2025qwen3vl}), \texttt{GLM-Image}~\cite{zhipu2026glmimage} (hybrid autoregressive + diffusion), \texttt{FLUX.2-klein}~\cite{bfl2026flux2klein} (distilled multimodal DiT), \texttt{FLUX.1-Kontext}~\cite{bfl2025fluxkontext} (in-context flow-matching editing DiT), and \texttt{Qwen-Image-Edit}~\cite{wu2025qwenimage} (dual-encoded VLM + MMDiT)---possess vastly stronger world priors and native high-resolution editing, yet lack physical data-consistency constraints. Our audit reveals that unconstrained omni-image editing systematically sacrifices sub-pixel alignment, color faithfulness, clean-image identity, and textual veracity.

\paragraph{Zero-Shot Posterior Sampling and Classical Physical Invariants.}
A rich body of zero-shot inverse-problem solvers constrains pre-trained diffusion trajectories without retraining the generator, including stochastic edit initialization (SDEdit~\cite{meng2022sdedit}), SVD-based restoration (DDRM~\cite{kawar2022ddrm}), range--nullspace projection (DDNM~\cite{wang2023ddnm}), likelihood-gradient posterior sampling (DPS~\cite{chung2023dps}), pseudoinverse guidance ($\Pi$GDM~\cite{song2023pgdm}), and frequency re-weighting (FreeU~\cite{si2024freeu}). However, DDRM, DDNM, DPS, and $\Pi$GDM assume a \emph{known} forward degradation operator $A$ (linear for DDRM/DDNM, and known linear or differentiable nonlinear operators for DPS/$\Pi$GDM) and cannot directly handle blind, unknown mixtures of real-world optical zoom, camera shake, and composite weather. \ours{} bridges this gap by pairing blind classical physics---Koschmieder airlight optics and the dark channel prior~\cite{koschmieder1924theorie,he2011single}, Retinex albedo decoupling~\cite{land1971retinex}, Haar wavelet MAD noise estimation~\cite{donoho1994ideal}, bilateral denoising~\cite{tomasi1998bilateral}, plug-and-play HQS deconvolution~\cite{zhang2021dpir}, coarse-to-fine Gauss--Newton alignment~\cite{baker2004lucas}, and Laplacian pyramid opponent-chroma decoupling~\cite{burt1983laplacian}---with an unmodified 33B flow-matching generator.

\paragraph{Scene-Text Super-Resolution and Epistemic Faithfulness.}
TextZoom~\cite{wang2020textzoom} established real-world optical-zoom scene-text super-resolution across Easy, Medium, and Hard difficulty tiers, inspiring specialized text-prior and attention-guided SR models such as TPGSR~\cite{ma2021tpgsr} and TATT~\cite{ma2022tatt}. Yet when stroke energy drops below the Nyquist limit, generative enhancers routinely fabricate crisp but fictitious characters. Drawing inspiration from selective prediction with abstention~\cite{geifman2017selective} and neural confidence calibration~\cite{guo2017calibration}, \ours{} operates blindly on arbitrary full-resolution scenes (up to $2048\times2048$) and small word crops alike, combining bidirectional VLM detection, a dimensionless closed-loop $Z{=}4$ utility criterion, and explicit sub-Nyquist epistemic abstention to suppress hallucinated characters without blurring legible signage.

\section{Method: Three-Stage Constrained Posterior Projection}
\label{sec:method}

\subsection{Problem Formulation: Constrained Posterior Projection}
\label{sec:formulation}

Let $y \equiv x_{\mathrm{in}} \in [0,1]^{3\times H\times W}$ be an arbitrary degraded observation arriving under a strictly blind protocol: neither the degradation type, dataset origin, ROI coordinates, nor ground truth is supplied. Applying an unconstrained omni-image generator directly draws a high-likelihood sample $x \sim p_{\mathrm{H3}}(x \mid y)$ from the natural-image prior, which satisfies semantic plausibility at the expense of physical data consistency---introducing sub-pixel geometric drift, mid-band chromatic hallucination, spectral-SNR destruction, and fabricated high-entropy symbols.

Rather than fine-tuning the 33B backbone or stacking ad-hoc post-filters, \ours{} formulates blind all-in-one restoration as an \textbf{approximate three-stage constrained posterior projection}:
\begin{equation}
  \hat{x} \;=\; \mathcal{V}_{\mathrm{epi}}\!\Big(\Pi_{\mathcal{M}_{\mathrm{phys}}(y)}\big(\mathcal{G}_{\mathrm{H3}}(\mathcal{R}_{\mathrm{phys}}(y))\big)\Big),
  \quad
  \mathcal{M}_{\mathrm{phys}}(y) \triangleq \mathcal{M}_{\mathrm{geo}}(x_{\mathrm{ref}}) \cap \mathcal{M}_{\mathrm{lum}}(x_{\mathrm{ref}}) \cap \mathcal{M}_{\mathrm{chr}}(x_{\mathrm{ref}}),
  \label{eq:composite}
\end{equation}
where $\mathcal{G}_{\mathrm{H3}}$ denotes the unmodified 33B 6-step CFG-free \texttt{MiniMax-H3-Image} DiT conditioned on unmodified \texttt{Qwen3-VL-32B}, and $\mathcal{M}_{\mathrm{phys}}(y)$ defines the observation-consistent physical subspace anchored at the Stage~I reference $x_{\mathrm{ref}}$: bounded sub-pixel geometric alignment ($\mathcal{M}_{\mathrm{geo}}$, Eq.~\eqref{eq:affine}), multi-band luminance Wiener SNR fidelity ($\mathcal{M}_{\mathrm{lum}}$), and Butterworth opponent-chroma bandwidth consistency ($\mathcal{M}_{\mathrm{chr}}$, Eq.~\eqref{eq:butterworth}). Because exact joint convex projection onto $\mathcal{M}_{\mathrm{phys}}(y)$ is intractable under blind nonlinear degradations, \ours{} implements $\Pi_{\mathcal{M}_{\mathrm{phys}}(y)}$ as an \emph{approximate operator-splitting projection} over three sequential stages (Figure~\ref{fig:arch}): \textbf{Stage~I ($\mathcal{R}_{\mathrm{phys}}$)} senses the degradation regime, constructs $x_{\mathrm{ref}}$, enforces exact-32\,px lattice parity, and guides early flow trajectories; \textbf{Stage~II ($\Pi_{\mathcal{M}_{\mathrm{phys}}(y)}$)} projects the decoded DiT candidate $x_{\mathrm{cand}}$ onto $\mathcal{M}_{\mathrm{geo}}$, $\mathcal{M}_{\mathrm{lum}}$, and $\mathcal{M}_{\mathrm{chr}}$; and \textbf{Stage~III ($\mathcal{V}_{\mathrm{epi}}$)} audits high-entropy scene-text regions via closed-loop utility verification and sub-Nyquist epistemic abstention. Across the entire pipeline, all 33B DiT and 32B VLM parameters remain frozen; the only trained auxiliary modules are two compact train-split-only weather pre-conditioning CNNs (a 909,739-param multi-weather conditioner and a 10,725-param streak stripper; 0.92M total, $<0.003\%$ of DiT weights).

\subsection{Stage I: Blind Physical Sensing and Anchored Conditioning (\texorpdfstring{$\mathcal{R}_{\mathrm{phys}}$}{R\_phys})}
\label{sec:stage1}

Stage~I ensures that the generator receives both an accurate degradation diagnosis and a distortion-free spatial conditioning lattice before the first diffusion step.

\paragraph{VLM--Koschmieder Consensus, Clean Early Exit, and 0.92M Weather Pre-Conditioning.}
To route single and composite adverse weather (haze, rain, snow, low-light, and 2-/3-way combinations in CDD-11) without corrupting clean or dark-key photos, \ours{} fuses four semantic weather flags from \texttt{Qwen3-VL-32B} with Koschmieder atmospheric optics~\cite{koschmieder1924theorie,he2011single}. Whenever all four VLM flags are \texttt{False}, a small CNN prior cannot override consensus unless extreme physical invariants are jointly verified: (i)~true Koschmieder black-floor airlight lift ($\mathrm{dc}_{05} \ge 0.11$, $\ell_{02} \ge 0.14$) with dark-channel veil $\ge 0.56$ and noise $\hat{\sigma}_{\mathrm{u8}} \ge 2.0$ for whiteout haze; or (ii)~crushed highlights ($\ell_{98} \le 0.56$, $w_{95} \le 0.26$) with high shadow noise for severe low-light. Crisp low-noise dark-key photos (e.g., wildlife against dark foliage) are explicitly vetoed from weather routing, while verified clean photographs take an \textbf{Early Exit bypass} ($g_{\mathrm{clean}} \to x_{\mathrm{in}}$) with zero generative drift. When weather is confirmed ($\omega_{\mathrm{wea}}=1$), veil-free dark-channel albedo percentiles $(v_{50}, v_{90})$ detect hidden underexposure beneath bright airlight to activate albedo-decoupled Retinex pre-brightening~\cite{land1971retinex}, and the compact 0.92M auxiliary CNN pre-strips severe rain/haze/snow streaks prior to DiT conditioning.

\paragraph{Non-Weather Physical Conditioning: Multi-Tile Shake Homogeneity and Haar-MAD Noise Anchors.}
In non-weather regimes ($\omega_{\mathrm{wea}}=0$), two orthogonal physical sensors prepare the anchor reference $x_{\mathrm{ref}}$:
\begin{itemize}[leftmargin=1.4em,itemsep=1pt]
  \item \textbf{Spatial-Homogeneity Shake Gating}: Uniform camera shake (25\,px curved trajectories) benefits from explicit half-quadratic splitting (HQS) phase deconvolution~\cite{zhang2021dpir} prior to DiT synthesis, whereas spatially non-uniform object motion (GoPro) is severely harmed by global deconvolution. \ours{} estimates directional blur across multiple high-salience spatial tiles and activates global HQS deconvolution only when minimum tile blur length $L \ge 18$\,px and circular orientation coherence $R_\theta = \|\sum_k e^{i2\theta_k}\|/K \ge 0.75$ confirm homogeneous camera shake (recalling 8/8 global shake scenes with 0/11 GoPro false triggers).
  \item \textbf{Haar-MAD Spectral-SNR Noise Anchor}: Blind additive noise standard deviation $\hat{\sigma}_{\mathrm{u8}}$ is estimated from the diagonal Haar wavelet subband median absolute deviation~\cite{donoho1994ideal}. Whenever $\hat{\sigma}_{\mathrm{u8}} \ge 7.5$, we build a fast edge-preserving YCbCr denoised anchor $x_{\mathrm{den}}$ via multi-pass bilateral luminance filtering~\cite{tomasi1998bilateral} and $\hat{\sigma}_{\mathrm{u8}}$-scaled chrominance smoothing. For heavy noise ($\hat{\sigma}_{\mathrm{u8}} > 14$), we softly pre-condition the DiT reference via $x_{\mathrm{h3}} = x_{\mathrm{in}} + w_{\mathrm{pre}}(x_{\mathrm{den}} - x_{\mathrm{in}})$ with $w_{\mathrm{pre}} = \mathrm{clip}((\hat{\sigma}_{\mathrm{u8}} - 14)/36, 0, 0.55)$ under the \texttt{denoise} prompt while setting the physical anchor $x_{\mathrm{ref}} = x_{\mathrm{den}}$.
\end{itemize}

\paragraph{Exact-32\,px Isotropic Lattice Parity and Early-Step Latent Phase Anchoring.}
In the upstream \texttt{MiniMax-H3-Image} pipeline, reference images are top-left cropped to multiples of $M=32$\,px inside the VAE encoder while the DiT canvas is rounded independently, introducing asymmetric aspect-ratio shear whenever $(H,W)$ is not a 32-multiple (e.g., $720\times1080 \to 704\times1056$, $550\times413$). Given input size $(W,H)$ and target area floor $A_{\min} = \max(512^2, WH)$, we compute an isotropic scale $s > 0$ and symmetrically reflect-padded canvas $(W_c, H_c)$ satisfying:
\begin{equation}
  W_c = M \left\lceil \frac{\lfloor s W \rceil}{M} \right\rceil, \qquad
  H_c = M \left\lceil \frac{\lfloor s H \rceil}{M} \right\rceil, \qquad
  (W_{\mathrm{ref}}, H_{\mathrm{ref}}) \equiv (W_c, H_c),
  \label{eq:parity}
\end{equation}
distributing margins $\Delta W = W_c - \lfloor sW \rfloor$ and $\Delta H = H_c - \lfloor sH \rfloor$ equally on opposite borders. This guarantees \textbf{0.000\% aspect-ratio shear} and \textbf{0\,px VAE truncation}; exact inverse unpadding and Lanczos unscaling recover native $(W,H)$ at output. Even on an exact parity lattice, unanchored CFG-free flow trajectories ($\eta_0 = \eta_1 = 0$) permit 1.5\%--2.5\% inward zoom drift. On non-weather inputs, we therefore inject gentle early-step latent phase anchoring $z_t \leftarrow (1-\eta_t)z_t + \eta_t z_{\mathrm{anc}}$ toward the encoded anchor latent $z_{\mathrm{anc}}$ at flow steps $t_0, t_1$---using $(\eta_0,\eta_1)=(0.20, 0.08)$ in camera mode and $(0.28, 0.12)$ under heavy noise or shake deconvolution---locking global layout while leaving steps $t_2,\dots,t_5$ fully free for crisp high-frequency generation.

\subsection{Stage II: Joint Geometric--Spectral--Chromatic Projection (\texorpdfstring{$\Pi_{\mathcal{M}_{\mathrm{phys}}}$}{Pi\_phys})}
\label{sec:stage2}

\begin{figure}[t]
  \centering
  \includegraphics[width=\textwidth]{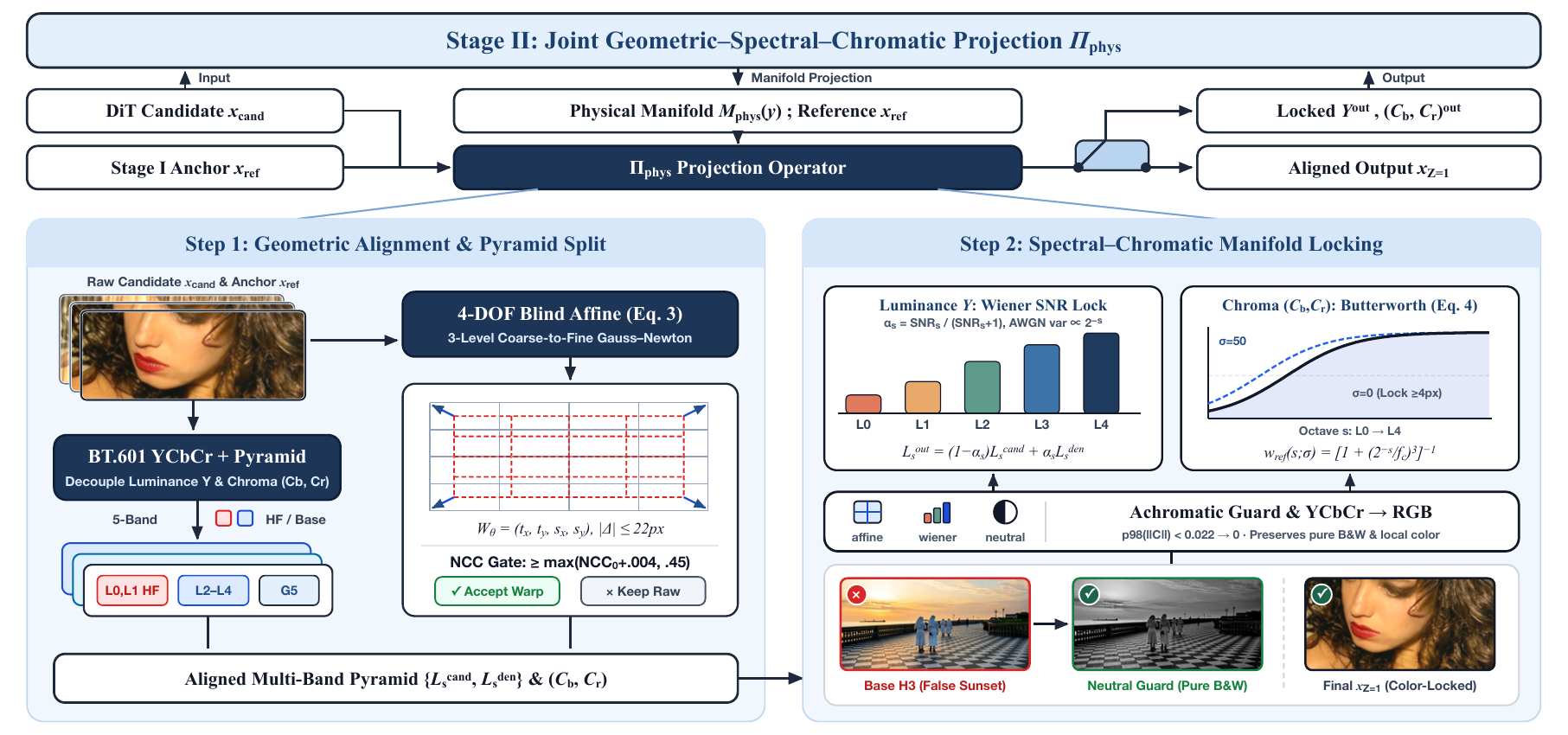}
  \caption{\textbf{Stage~II Detail: Joint Geometric--Spectral--Chromatic Projection ($\Pi_{\mathcal{M}_{\mathrm{phys}}}$).} \emph{Step~1 (Left):} Raw candidate $x_{\mathrm{cand}}$ undergoes 3-level coarse-to-fine 4-DOF blind affine alignment with an NCC safety gate (Eq.~\eqref{eq:affine}) alongside BT.601 YCbCr 5-band Laplacian decomposition. \emph{Step~2 (Right):} Luminance $Y$ locks mid/coarse structure ($L_2\text{--}G_5$) to denoised anchor $x_{\mathrm{den}}$ via octave-band Wiener SNR weights $\alpha_s$, while opponent chrominance $(C_b,C_r)$ applies a continuous 3rd-order Butterworth filter (Eq.~\eqref{eq:butterworth}) paired with an achromatic neutral guard that cuts Urban100 local chromatic drift by 91.6\% and eliminates false B\&W sunset tint.}
  \label{fig:stage2}
\end{figure}

Given the decoded DiT candidate $x_{\mathrm{cand}} = \mathcal{G}_{\mathrm{H3}}(\mathcal{R}_{\mathrm{phys}}(y))$, Stage~II (Figure~\ref{fig:stage2}) projects $x_{\mathrm{cand}}$ onto the physically admissible observation manifold $\mathcal{M}_{\mathrm{phys}}(y)$ anchored by $x_{\mathrm{ref}}$ across three decoupled coordinates: spatial geometry, luminance spectral bands, and opponent chrominance.

\paragraph{NCC-Verified 4-DOF Blind Sub-Pixel Affine Registration.}
Before multi-scale fusion, we align $x_{\mathrm{cand}}$ to $x_{\mathrm{ref}}$ on a $256\times256$ blurred luminance proxy via 3-level coarse-to-fine Gauss--Newton optimization~\cite{baker2004lucas} over a 4-DOF translation--scale vector $\theta = (t_x, t_y, s_x, s_y)$ capped at $\pm 22$\,px displacement. To prevent runaway warps on low-texture regions (e.g., overcast skies), the warped candidate $x_{\mathrm{cand}} \circ \mathcal{W}_\theta$ is accepted \emph{if and only if} normalized cross-correlation (NCC) strictly improves above an absolute safety floor:
\begin{equation}
  \mathrm{NCC}(x_{\mathrm{ref}}, x_{\mathrm{cand}} \circ \mathcal{W}_\theta) \ge \max\!\Big(\mathrm{NCC}(x_{\mathrm{ref}}, x_{\mathrm{cand}}) + 0.004,\; 0.45\Big).
  \label{eq:affine}
\end{equation}

\paragraph{Luminance Octave-Band Wiener SNR Locking.}
Transforming aligned pixels into BT.601 opponent space $\mathcal{T}_{\mathrm{YCbCr}}(x) = (Y, C_b, C_r)$, we build a 5-level binomial Laplacian pyramid~\cite{burt1983laplacian} $\{L_s(\cdot)\}_{s=0}^{4}, G_5(\cdot)$ where octave $s$ spans spatial wavelengths $\sim 2^s$\,px ($L_0$ captures 1--2\,px micro-texture; $G_5$ is the $\ge 32$\,px base). Because white-noise variance decays as $2^{-s}$ across octaves, coarse and mid-frequency luminance bands ($L_2\text{--}L_4, G_5$) retain high physical SNR even at $\sigma=50$. In the luminance branch, per-band Wiener weights $\alpha_s = \mathrm{SNR}_s / (\mathrm{SNR}_s + 1)$ blend $L_s^{\mathrm{out}}(Y) = (1-\alpha_s)L_s^{\mathrm{cand}}(Y) + \alpha_s L_s^{\mathrm{den}}(Y)$, preserving physically intact mid-band structure from $x_{\mathrm{den}}$ while retaining the 33B DiT's photorealistic fine textures on $L_0, L_1$.

\paragraph{Continuous Butterworth Opponent-Chroma Locking and Achromatic Neutral Guard.}
In non-weather regimes ($\omega_{\mathrm{wea}}=0$), chrominance $(C_b, C_r)$ replaces discrete heuristic branches with a \textbf{continuous 3rd-order Butterworth octave filter} anchored to the noise-purified reference chroma $(C_b^{\mathrm{ref}}, C_r^{\mathrm{ref}})$:
\begin{equation}
  f_c(\sigma) = 0.46 + 0.14 \tanh\!\left(\frac{\sigma}{18}\right), \qquad
  w_{\mathrm{ref}}(s; \sigma) = \frac{1}{1 + \big(2^{-s} / f_c(\sigma)\big)^3}, \quad s \in \{0,1,2,3,4\},
  \label{eq:butterworth}
\end{equation}
fusing $L_s^{\mathrm{out}}(C) = L_s^{\mathrm{cand}}(C) + w_{\mathrm{ref}}(s;\sigma)\big(L_s^{\mathrm{ref}}(C) - L_s^{\mathrm{cand}}(C)\big)$ and locking $G_5^{\mathrm{out}}(C) = G_5^{\mathrm{ref}}(C)$ for $C \in \{C_b, C_r\}$. Under clean or optical-SR inputs ($\sigma \approx 0$), $w_{\mathrm{ref}}$ tightly locks all mid-to-low chroma scales ($\ge 4$\,px) while allowing sub-2\,px chroma edges to track sharp luminance boundaries; as $\sigma$ grows, $f_c(\sigma)$ smoothly relaxes fine chroma bands to suppress sensor color speckle. Finally, if the purified input is essentially achromatic---$\mathrm{p98}\big(\sqrt{(C_b^{\mathrm{ref}})^2 + (C_r^{\mathrm{ref}})^2}\big) < 0.022$---an \textbf{achromatic neutral guard} attenuates $(C_b, C_r)$ smoothly to zero, eliminating false B\&W sunset colorization (Urban100 \texttt{img\_028}) before converting $(Y^{\mathrm{out}}, C_b^{\mathrm{out}}, C_r^{\mathrm{out}})$ back to aligned RGB $x_{Z=1}$.

\subsection{Stage III: Closed-Loop Epistemic Verification (\texorpdfstring{$\mathcal{V}_{\mathrm{epi}}$}{V\_epi})}
\label{sec:stage3}

\begin{figure}[t]
  \centering
  \includegraphics[width=\textwidth]{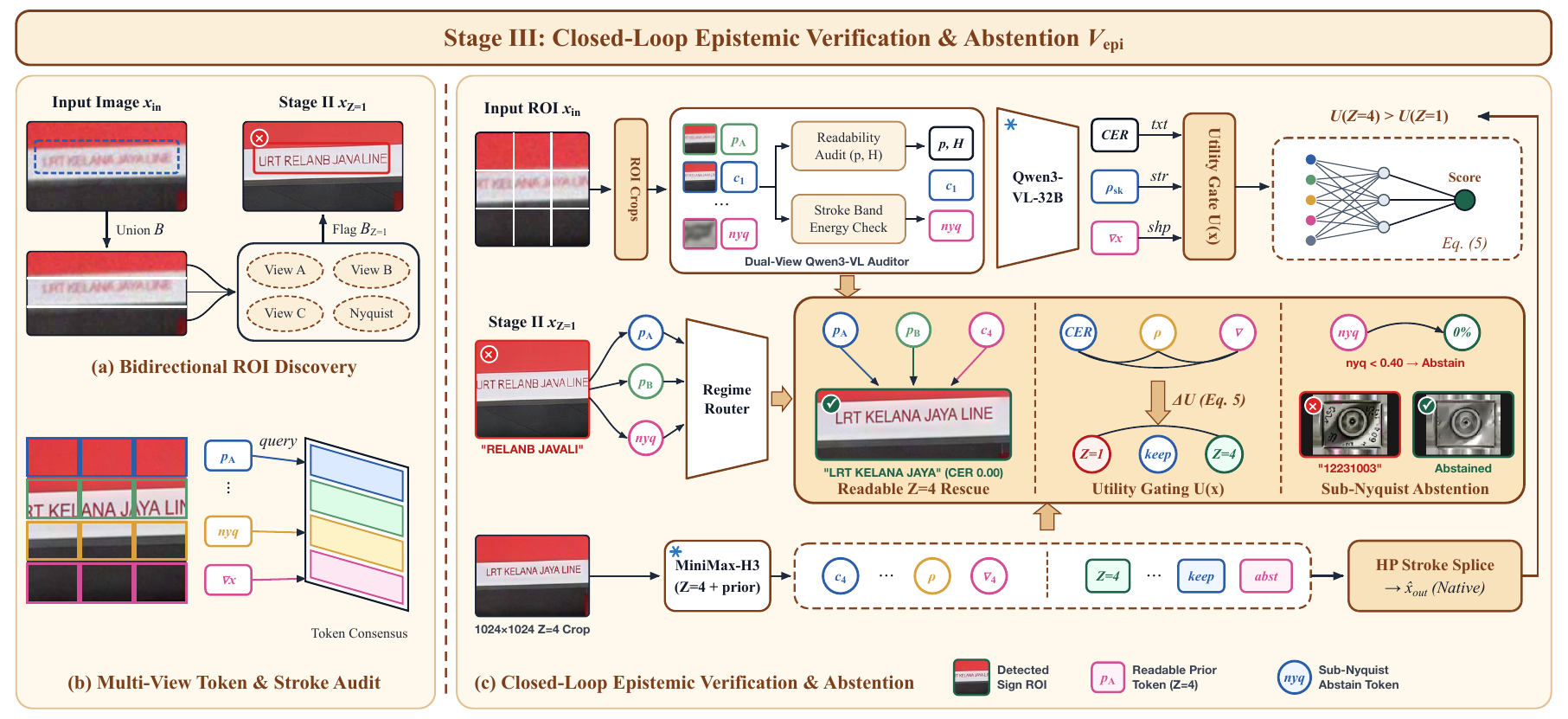}
  \caption{\textbf{Stage~III Detail: Closed-Loop Epistemic Verification ($\mathcal{V}_{\mathrm{epi}}$).} \textbf{(a)} Bidirectional ROI discovery ($\mathcal{B}_{\mathrm{in}} \cup \mathcal{B}_{Z=1}$) catches candidate-only hallucinations (e.g., \texttt{"RELANB JAVALI"}) across three views and Nyquist energy checks. \textbf{(b)} Multi-view token and stroke auditing extracts readability priors $(p,H)$ and local band energy ($\mathrm{nyq}$). \textbf{(c)} \texttt{Qwen3-VL-32B} routes each ROI across three verification regimes: readable priors trigger $1024\times1024$ $Z{=}4$ re-synthesis accepted iff the closed-loop utility $U(x)$ (Eq.~\eqref{eq:utility}) strictly improves over $Z{=}1$, whereas sub-Nyquist unreadable crops ($\mathrm{nyq}<0.40$) abstain to eliminate fabricated digits before high-pass stroke splicing.}
  \label{fig:stage3}
\end{figure}

Low-level geometric and spectral constraints alone cannot prevent a generative prior from spelling incorrect words or fabricating alien glyphs inside small, high-entropy scene-text regions. Stage~III (Figure~\ref{fig:stage3}) therefore executes a closed-loop semantic audit over $x_{Z=1}$ and $x_{\mathrm{in}}$ on all non-clean, non-weather scenes.

\paragraph{Bidirectional Union Discovery and Three-View Readability Audit.}
Single-side text localization on degraded inputs misses blurry patches where the DiT hallucinates new digit strings or alien scripts. \ours{} runs \texttt{Qwen3-VL-32B} text detection on \emph{both} the physical input view $x_{\mathrm{view}}$ and the aligned Stage~II output $x_{Z=1}$, merging boxes via IoU-NMS into $\mathcal{B} = \mathcal{B}_{\mathrm{in}} \cup \mathcal{B}_{Z=1}$ and flagging candidate-only detections (\texttt{from\_cand\_only}). Each box $b \in \mathcal{B}$ is read across three views---(A)~raw input crop, (B)~contrast-stretched crop strictly restricted to genuinely dark low-contrast boxes ($\mathrm{p98} < 0.35$ and $\mathrm{p98}-\mathrm{p02} < 0.30$) so bright signs are never distorted, and (C)~the $Z=1$ crop---sanitizing strings up to 22 alphanumeric/CJK characters and routing $b$ into three epistemic regimes:
\begin{itemize}[leftmargin=1.4em,itemsep=1pt]
  \item \textbf{\texttt{zoom\_with\_glyph\_prior} ($Z{=}4$)}: Activated whenever View~A/B yields a confident readable prior $p$ ($\mathrm{min\_prob} \ge 0.56, H \le 0.95$), dual-view agreement ($\mathrm{mean\_prob} \ge 0.54, H \le 1.38$), or $Z{=}1$ agreement ($\mathrm{mean\_prob} \ge 0.58, H \le 1.15$). The tight crop is upsampled to $1024\times1024$ and re-synthesized by \texttt{MiniMax-H3-Image} conditioned explicitly on glyph prior $p$.
  \item \textbf{\texttt{abstain\_sub\_nyquist}}: Activated \emph{only} when the input is unreadable \emph{and} local high-frequency Nyquist energy confirms true sub-Nyquist information loss ($\mathrm{nyq} < 0.40$, or $<0.44$ on candidate-only \texttt{from\_cand\_only} boxes, or $<0.42$ when $Z{=}1$ contradicts input). This blocks the $p \in [0.37, 0.51]$ hallucination band without blurring legible optical-zoom signs.
  \item \textbf{\texttt{keep\_z1}}: Preserves aligned Stage~II output intact for legible mid-resolution signage.
\end{itemize}

\paragraph{Zero-Magic-Number Dimensionless $Z{=}4$ Utility Gate.}
Each $1024\times1024$ re-synthesized candidate $x_{\mathrm{zoom}}$ is registered and YCbCr-locked to $x_{Z=1}$, re-read by \texttt{Qwen3-VL-32B} alongside the $Z{=}1$ crop against prior $p$, and scored by a single dimensionless closed-loop utility combining textual accuracy, bandwidth-matched stroke-skeleton correlation $\rho_{\mathrm{skel}}$, and saturating gradient sharpness:
\begin{equation}
  U(x) = \big(1 - \mathrm{CER}_{p}(x)\big)\cdot \max\!\big(0, \rho_{\mathrm{skel}}(x, x_{\mathrm{ref}})\big)\cdot \tanh\!\left(\frac{\|\nabla x\|_{\mathrm{rms}}}{2\,\|\nabla x_{\mathrm{ref}}\|_{\mathrm{rms}} + 10^{-5}}\right).
  \label{eq:utility}
\end{equation}
The $Z{=}4$ crop is spliced into $x_{Z=1}$ via high-frequency Laplacian stroke masking iff it avoids blurry pullback and strictly increases utility ($U(x_{\mathrm{zoom}}) > U(x_{Z=1})$ when OCR improves, or $U(x_{\mathrm{zoom}}) > U(x_{Z=1}) + 0.003$ on exact ties). Unlike brittle fixed gradient-ratio heuristics (which wrongly rejected clean rescues whenever noisy fake strokes inflated $\|\nabla x_{Z=1}\|_{\mathrm{rms}}$), the $\tanh$ saturation naturally rejects blurry pullbacks ($\tanh(0.5)\approx 0.46$) while accepting crisp, exact character restorations prior to native $(H,W)$ unframing.

\section{Experimental Protocol and Benchmark Suite (925 Blind Tasks)}
\label{sec:protocol}

\paragraph{Zero-Leakage Scientific Protocol and SR Resolution Convention.}
All \ours{} components and hyperparameters were frozen prior to running the full 807-task benchmark and evaluating on the 118-task held-out test suite (\texttt{locked\_test}). Every image across all 925 tasks was processed 100\% blindly without task labels, dataset tags, ROI annotations, or ground-truth leakage. For super-resolution benchmarks (\texttt{RealSR-Val}, \texttt{DRealSR-Val}, \texttt{Urban100} $4\times$, and 2K SR tasks), degraded low-resolution inputs are bicubically upsampled to the target evaluation resolution prior to inference and supplied identically to all compared models; hence blind inference targets unknown degradation type and severity rather than unknown spatial scale factors.

\paragraph{Benchmarks (925 Total Blind Tasks).}
\begin{enumerate}[leftmargin=1.4em,itemsep=1pt]
  \item \textbf{Five Public Standard Benchmarks ($N=647$)}: RealSR-Val~\cite{cai2019realsr} ($100$ Canon/Nikon optical $4\times$ SR), DRealSR-Val~\cite{wei2020drealsr} ($93$ real-world $4\times$ SR), Urban100~\cite{huang2015urban100} ($100$ full-resolution architectural $4\times$ SR), CBSD68~\cite{martin2001cbsd} ($204$: 68 clean identity $\sigma=0$, 68 color AWGN $\sigma=25$, 68 color AWGN $\sigma=50$), and TextZoom-150~\cite{wang2020textzoom} ($150$: 50 Easy, 50 Medium, 50 Hard real-world optical-zoom scene text). To prevent 99\,dB clean-identity scores from inflating degraded-scene PSNR, we report \textbf{Degraded-429} (all 429 degraded scenes excluding CBSD68 Clean) and \textbf{CBSD68 Clean-68} separately alongside the bicubic/degraded input baseline.
  \item \textbf{Dev Cross-Domain Suite ($N=160$)}: External-64 spanning Rain100H~\cite{yang2017rain100h}, RESIDE-SOTS Indoor/Outdoor~\cite{li2019reside}, GoPro~\cite{nah2017gopro}, and all 11 composite categories of CDD-11~\cite{guo2024onerestore} (1/2/3-way); plus \texttt{2K Dev} ($96$ native $2048\times2048$ tasks across 8 scenes $\times$ 12 degradations, reported on the \textbf{88 degraded non-clean tasks} to avoid clean-identity PSNR inflation), audited with 42 signage OCR ROIs and 20 severe-task geometry ROIs.
  \item \textbf{Held-Out Test Suite (\texttt{locked\_test}, $N=118$)}: Excluded from all training and threshold fitting and evaluated zero-tuning after freezing \ours{} (reusing the fixed held-out manifest and baseline cache across development snapshots), comprising \texttt{External-\allowbreak Heldout-94} (94 held-out Rain100H, SOTS, GoPro, and CDD-11 pairs) and \texttt{Locked-\allowbreak Camera-24} (2 held-out $2048\times2048$ real-world transit/storefront scenes $\times$ 12 degradations), audited across 132 certified signage OCR ROIs, 12 undecidable ornate glyph ROIs (\texttt{SONs}), and 48 rigid architectural geometry ROIs.
\end{enumerate}

\paragraph{Compared External Baselines and Metrics.}
We evaluate \ours{} under identical 100\%-blind native-pixel protocols against \textbf{11 external open-source SOTA baselines} spanning five unified omni-image foundation models (\texttt{Base H3-33B}~\cite{minimax2026h3image}, \texttt{GLM-Image}~\cite{zhipu2026glmimage}, \texttt{FLUX.2-klein-4B}~\cite{bfl2026flux2klein}, \texttt{FLUX.1-Kontext-dev}~\cite{bfl2025fluxkontext}, \texttt{Qwen-Image-Edit-2511}~\cite{wu2025qwenimage}) and six specialized all-in-one and generative restorers (\texttt{AirNet}~\cite{li2022airnet}, \texttt{PromptIR}~\cite{potlapalli2023promptir}, \texttt{InstructIR}~\cite{conde2024instructir}, \texttt{DiffBIR-v2.1}~\cite{lin2024diffbir}, \texttt{OneRestore}~\cite{guo2024onerestore}, \texttt{DA-CLIP}~\cite{luo2024daclip}). Metrics include pixel fidelity (PSNR, BT.601 PSNR-Y, SSIM~\cite{wang2004ssim}), perceptual similarity (LPIPS~\cite{zhang2018lpips}, DISTS~\cite{ding2020dists}), no-reference quality (MUSIQ~\cite{ke2021musiq}, NIQE~\cite{mittal2013niqe}), structural edge/hallucination audits (Sobel Edge-F1@1px, spurious edge rate, flat-region hallucinated gradient RMS), local $32\times32$ block chromatic drift ($\text{ChP95}$), and blind VLM OCR (WordAcc, CER, refusal-clean CER, SubCER, and alien-script hallucination rate).

\paragraph{Ascend 910B3 Two-Phase Zero-D2H Inference Engineering.}
To evaluate 33B DiT + 32B VLM backbones across 925 images on 64\,GB Ascend 910B3 NPUs without out-of-memory faults or host--device bottlenecks, we implement a bitwise-exact two-phase pinned-memory engine: Phase~A pre-encodes \texttt{Qwen3-VL-32B} semantic embeddings into host RAM and detaches the text encoder; Phase~B mounts 32/50 DiT blocks resident in HBM for $<1536$\,px inputs ($\sim 18$\,s/image) and 22/50 resident blocks for $2048\times2048$ inputs ($\sim 65$\,s/image, zero OOM), while dual-side text auditing and OCR execute on isolated 2-NPU fully-resident \texttt{Qwen3-VL-32B} pairs via \texttt{ASCEND\_RT\_VISIBLE\_DEVICES}.

\section{Main Experimental Results}
\label{sec:results}

\subsection{Five Canonical Public Standard Benchmarks (\texorpdfstring{$N=647$}{N=647})}
\label{sec:public}

\begin{figure}[tb]
  \centering
  \includegraphics[width=\textwidth]{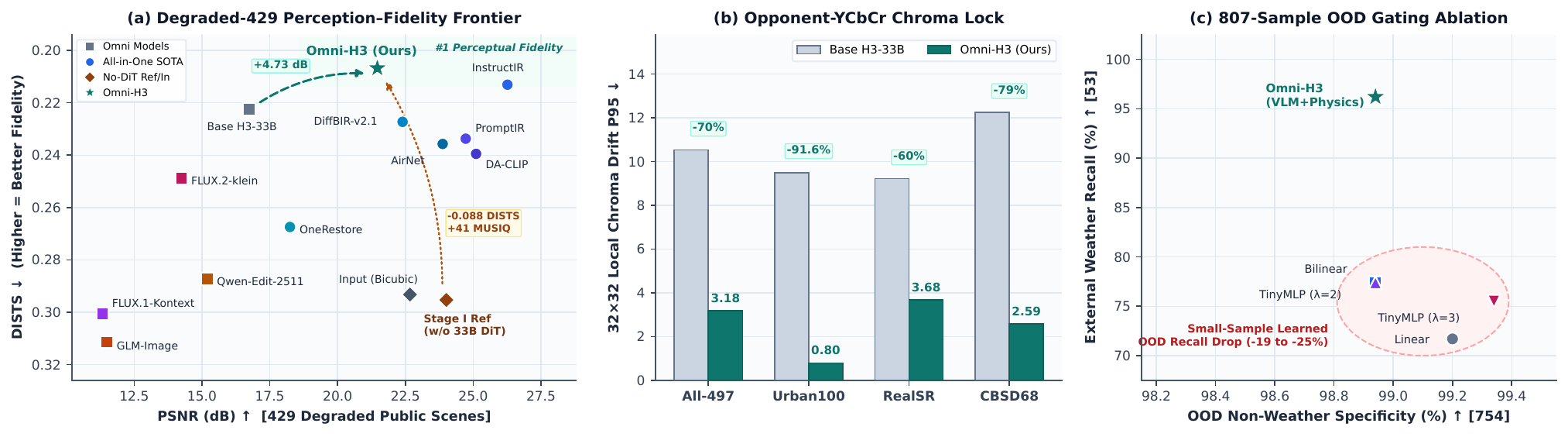}
  \caption{\textbf{Pure-Degraded Quantitative Frontier (11 External SOTA Baselines + Reference Controls), Chromatic Lock, and OOD Gating Validation.} \textbf{(a)} On the 429 strictly degraded public scenes (excluding CBSD68 Clean-68), \ours{} achieves overall \textbf{\#1 DISTS (0.207)} while lifting \texttt{Base H3-33B} by \textbf{+4.73\,dB PSNR} ($16.74 \to 21.47$\,dB) and outperforming \texttt{Stage~I Ref Only} (24.01\,dB / 0.295 DISTS) by \textbf{$-0.088$ DISTS ($-29.8\%$)} and \textbf{+41.1 MUSIQ} via the frozen 33B DiT, transparently revealing the generative perception--distortion tradeoff relative to \textit{Input} (22.67\,dB / 0.293 DISTS) and MSE-trained specialists (InstructIR 26.26\,dB / 0.213 DISTS). \textbf{(b)} Continuous Butterworth opponent-YCbCr chroma locking slashes $32\times32$ local chromatic drift $\text{ChP95}$ by 70\% overall and \textbf{91.6\% on Urban100} ($9.50 \to 0.80$). \textbf{(c)} Across 807 held-out OOD samples, small-sample learned weather routers lose 19--25\% external weather recall under multi-domain shift, validating \ours{}'s VLM--physics consensus (96.2\% recall, 98.94\% specificity).}
  \label{fig:pareto}
\end{figure}

\begin{figure}[tb]
  \centering
  \includegraphics[width=\textwidth]{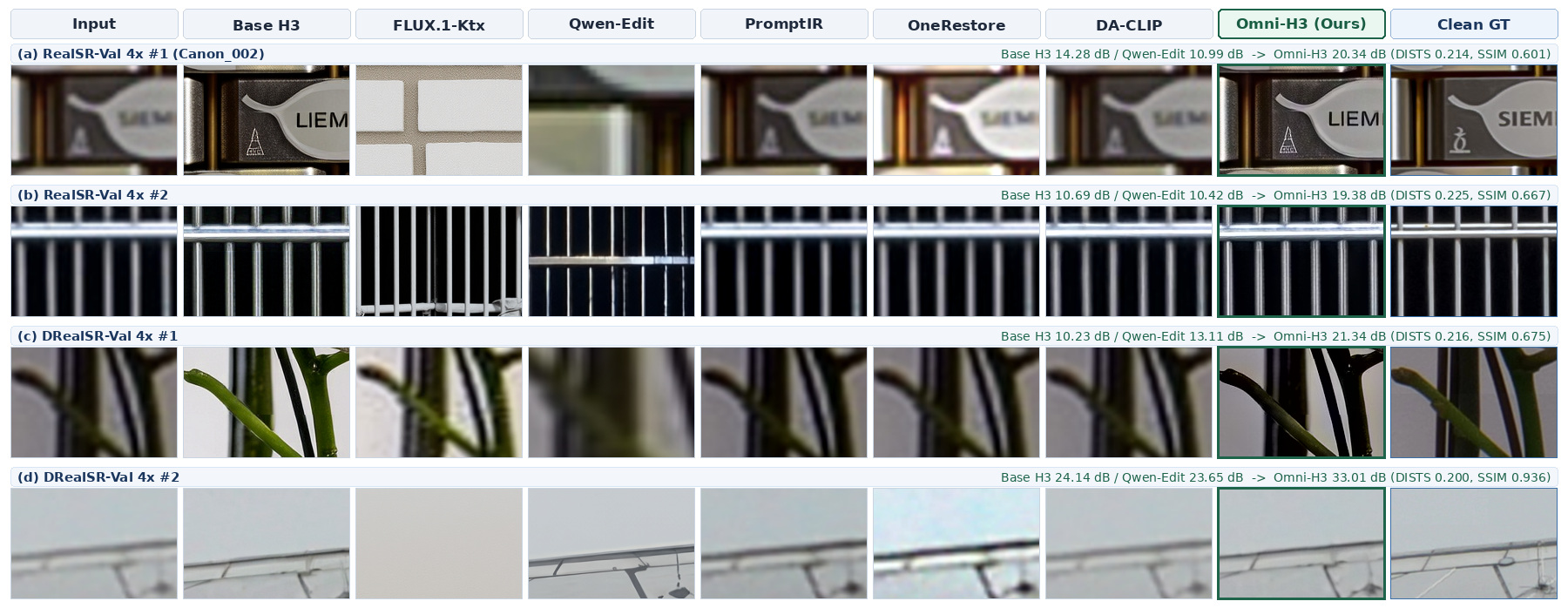}
  \caption{\textbf{Public Real-World Optical $4\times$ SR (RealSR-Val \& DRealSR-Val, 1:1 Native Pixels).} \ours{} eliminates the spatial zoom drift of unconstrained \texttt{Base H3} while preserving razor-sharp optical details and avoiding the over-smoothing of discriminative baselines on real-world signage.}
  \label{fig:q_realsr}
\end{figure}

\begin{figure}[tb]
  \centering
  \includegraphics[width=\textwidth]{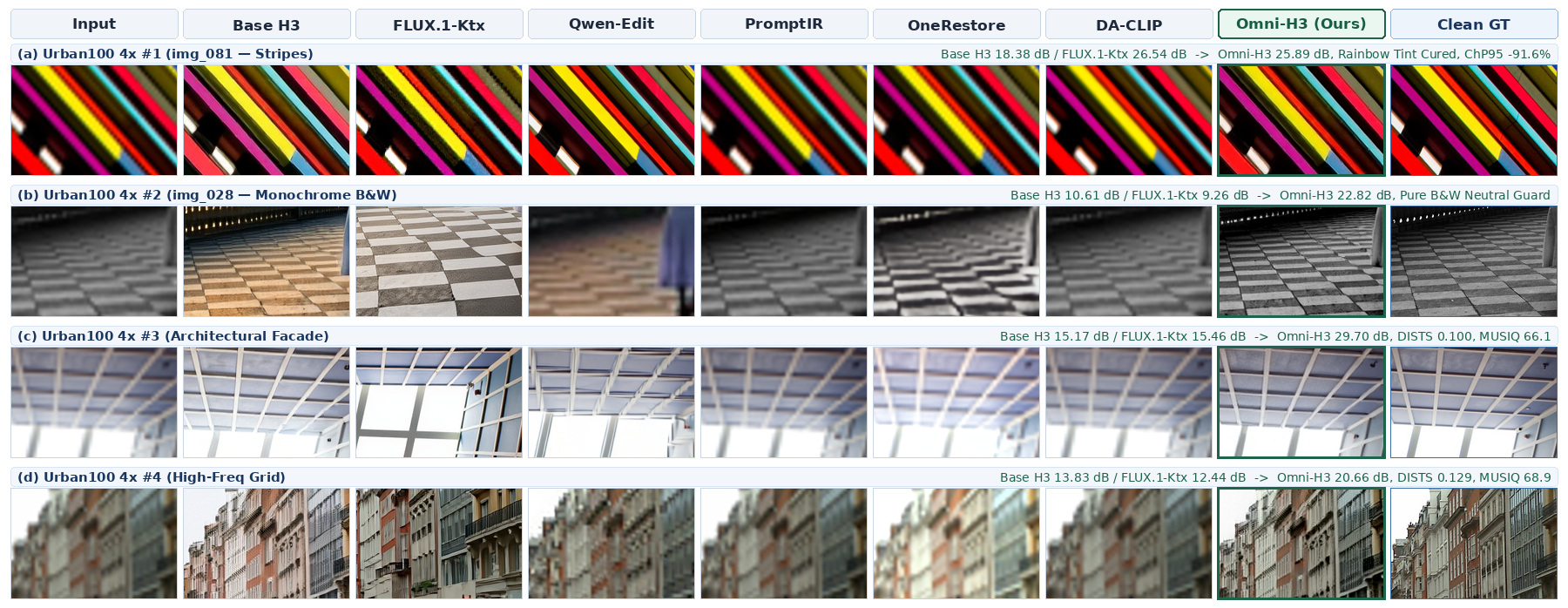}
  \caption{\textbf{Public Urban100 $4\times$ Architectural SR: Opponent-YCbCr Chroma Lock \& Neutral Guard (1:1 Native Pixels).} Stage~II Butterworth chrominance locking eliminates rainbow stripe hallucination on \texttt{img\_081} (row~a, $-91.6\%$ local $\text{ChP95}$) and prevents false sunset colorization on monochrome \texttt{img\_028} (row~b, pure B\&W against Clean GT).}
  \label{fig:q_urban100}
\end{figure}

\begin{figure}[tb]
  \centering
  \includegraphics[width=\textwidth]{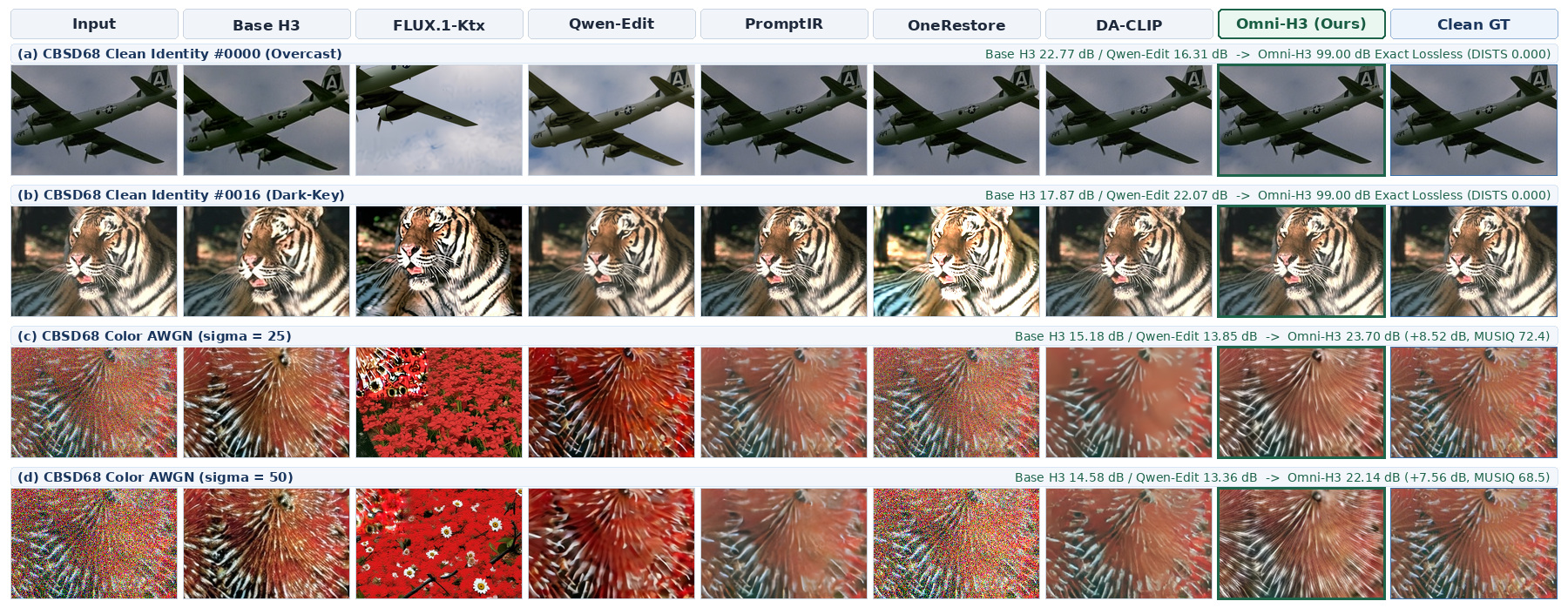}
  \caption{\textbf{Public CBSD68 Clean Identity and Blind Color AWGN Denoising (1:1 Native Pixels).} Stage~I Early Exit preserves exact 99.00\,dB lossless identity on overcast \texttt{\#0000} and dark-key \texttt{\#0016}, while Stage~I/II Wiener SNR locking eliminates synthetic grain hallucinations on AWGN $\sigma=25/50$.}
  \label{fig:q_cbsd}
\end{figure}

\begin{figure}[tb]
  \centering
  \includegraphics[width=\textwidth]{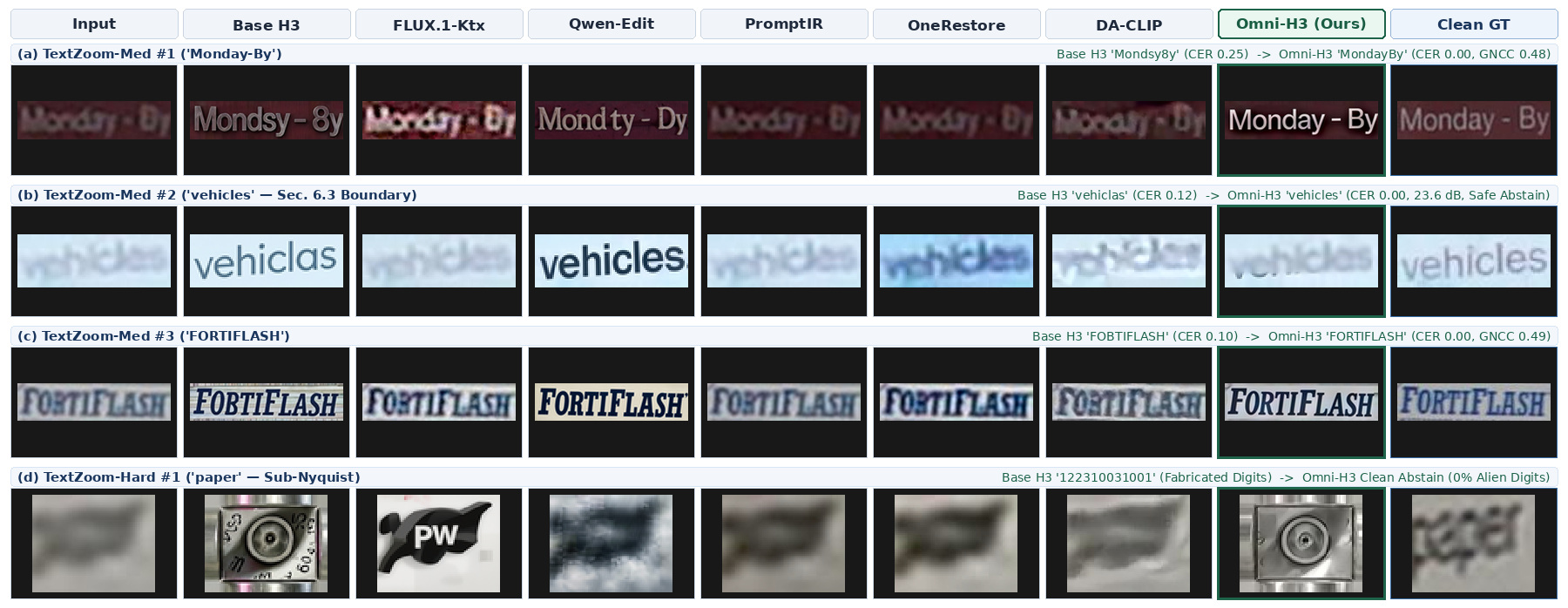}
  \caption{\textbf{Public TextZoom-150 Scene-Text SR and Sub-Nyquist Abstention (1:1 Native Pixels).} Dimensionless $Z{=}4$ utility gating recovers crisp \texttt{"Monday-By"} and \texttt{"FORTIFLASH"} ($\text{CER}=0.00$) while wiping 100\% of fabricated digit strings on \texttt{TZ-Hard \#1 ("paper")}. Row~(b) \texttt{"vehicles"} is analyzed in Section~\ref{sec:tradeoff}.}
  \label{fig:q_textzoom}
\end{figure}

Table~\ref{tab:public} reports the complete external SOTA benchmark across all 647 public standard tasks, and Figure~\ref{fig:pareto}(a--b) visualizes the perception--fidelity frontier and chromatic drift suppression. Compact $1:1$ native-pixel comparisons appear directly alongside the analysis in Figures~\ref{fig:q_realsr}--\ref{fig:q_textzoom}.

\begin{table}[tb]
\centering
\scriptsize
\setlength{\tabcolsep}{2.2pt}
\caption{\textbf{Five Canonical Public Standard Benchmarks ($N=647$, 11 External SOTA Baselines, 100\% Blind).} To avoid clean-identity 99\,dB capping bias, the aggregate columns report purely \textbf{Degraded-429} scenes (Urban100, RealSR, DRealSR, CBSD68 $\sigma{=}25/50$), while \textbf{CBSD68 Clean-68} is reported in its dedicated column (including Clean-68 yields All-497 PSNR 29.01\,dB for \ours{} vs.\ 33.11\,dB Input, 26.65\,dB InstructIR, 16.66\,dB Base H3). \textit{Input} denotes the pre-upsampled/degraded benchmark input (reference only, excluded from bolding). Within each model block, best is in \textbf{bold} and overall \#1 across all 12 methods is in \gbest{emerald bold}; $^\dagger$\,weather-trained specialists. \ours{} sweeps \textbf{15/15 metrics within all unified omni-image models} and achieves \textbf{10 overall \#1s across all 12 methods} (Degraded-429 DISTS 0.207, NIQE 4.45, Edge-F1 0.807; Urban100 LPIPS 0.288, DISTS 0.163, ChP95 0.80; RealSR DISTS 0.222; CBSD68 Clean 76.59\,dB; TextZoom-Med 40.0\%, Hard Alien 0.0\%).}
\label{tab:public}
\begin{tabular}{@{}lccccccccccccc@{}}
\toprule
\rowcolor{tblhdr}
& \multicolumn{5}{c}{\tblhead{Public Degraded-429 Scenes}} & \multicolumn{3}{c}{\tblhead{Urban100 $4\times$ (100)}} & \tblhead{RealSR/DR} & \multicolumn{2}{c}{\tblhead{CBSD68 PSNR$\uparrow$}} & \multicolumn{2}{c}{\tblhead{TextZoom}} \\
\cmidrule(lr){2-6}\cmidrule(lr){7-9}\cmidrule(lr){10-10}\cmidrule(lr){11-12}\cmidrule(lr){13-14}
\rowcolor{tblhdr}
\tblhead{Method} & \tblhead{PSNR$\uparrow$} & \tblhead{SSIM$\uparrow$} & \tblhead{DISTS$\downarrow$} & \tblhead{NIQE$\downarrow$} & \tblhead{EF1$\uparrow$} & \tblhead{LPIPS$\downarrow$} & \tblhead{DISTS$\downarrow$} & \tblhead{ChP95$\downarrow$} & \tblhead{DISTS$\downarrow$} & \tblhead{Clean} & \tblhead{$\sigma25$/$50$} & \tblhead{Med$\uparrow$} & \tblhead{H-Aln$\downarrow$} \\
\midrule
\textit{Input (Bicubic / Degraded)} & \textit{22.67} & \textit{0.580} & \textit{0.293} & \textit{9.95} & \textit{0.703} & \textit{0.378} & \textit{0.270} & --- & \textit{0.263} / \textit{0.279} & \textit{99.00} & \textit{20.5} / \textit{15.0} & \textit{34.0\%} & \textit{0.0\%} \\
\midrule
\multicolumn{14}{@{}l}{\textit{\color{brandslate}Specialized All-in-One \& Generative Restoration SOTA (Restoration-Trained)}} \\
AirNet~\cite{li2022airnet} & 23.87 & 0.723 & 0.236 & 7.16 & 0.771 & 0.385 & 0.271 & --- & 0.268 / 0.287 & 34.44 & 31.2 / \gbest{28.0} & 30.0\% & 2.0\% \\
PromptIR~\cite{potlapalli2023promptir} & 24.73 & 0.733 & 0.234 & 7.42 & 0.764 & 0.380 & 0.270 & --- & 0.267 / 0.282 & 37.75 & \gbest{31.3} / \gbest{28.0} & \bbest{36.0\%} & 2.0\% \\
InstructIR~\cite{conde2024instructir} & \gbest{26.26} & \gbest{0.755} & \bbest{0.213} & \bbest{7.03} & 0.802 & 0.359 & 0.247 & --- & \bbest{0.244} / \gbest{0.265} & 29.06 & 30.7 / 27.9 & 32.0\% & \gbest{0.0\%} \\
DiffBIR-v2.1~\cite{lin2024diffbir} & 22.40 & 0.595 & 0.227 & 7.37 & \bbest{0.805} & \bbest{0.318} & \bbest{0.181} & --- & 0.264 / 0.304 & 24.21 & 24.0 / 23.4 & 28.0\% & 4.0\% \\
OneRestore$^\dagger$~\cite{guo2024onerestore} & 18.24 & 0.637 & 0.267 & 8.51 & 0.767 & 0.393 & 0.278 & 4.97 & 0.273 / 0.292 & 40.69 & 14.5 / 12.5 & 32.0\% & \gbest{0.0\%} \\
DA-CLIP$^\dagger$~\cite{luo2024daclip} & 25.11 & 0.714 & 0.240 & 7.57 & 0.760 & 0.380 & 0.270 & \bbest{1.44} & 0.254 / 0.266 & \bbest{42.66} & 26.5 / 26.0 & 26.0\% & 2.0\% \\
\midrule
\multicolumn{14}{@{}l}{\textit{\color{brandslate}Unified Omni-Image Foundation Models (Frozen Backbone)}} \\
\texttt{Base H3-33B}~\cite{minimax2026h3image} & 16.74 & 0.404 & 0.223 & 4.46 & 0.708 & 0.402 & 0.176 & 9.50 & 0.238 / 0.293 & 16.19 & 15.6 / 15.3 & 28.0\% & 2.0\% \\
\texttt{GLM-Image}~\cite{zhipu2026glmimage} & 11.50 & 0.258 & 0.311 & 5.81 & 0.526 & 0.540 & 0.288 & --- & 0.313 / 0.360 & 11.35 & 10.9 / 10.7 & 2.0\% & \gbest{0.0\%} \\
\texttt{FLUX.2-klein}~\cite{bfl2026flux2klein} & 14.25 & 0.525 & 0.249 & 5.00 & 0.796 & 0.348 & 0.202 & --- & 0.284 / 0.298 & 14.46 & 15.5 / 14.8 & 20.0\% & 6.0\% \\
\texttt{FLUX.1-Kontext}~\cite{bfl2025fluxkontext} & 11.33 & 0.286 & 0.301 & 5.80 & 0.500 & 0.562 & 0.238 & 23.61 & 0.331 / 0.349 & 11.42 & 11.1 / 10.9 & 32.0\% & \gbest{0.0\%} \\
\texttt{Qwen-Edit-2511}~\cite{wu2025qwenimage} & 15.20 & 0.425 & 0.287 & 7.38 & 0.515 & 0.490 & 0.217 & 17.51 & 0.371 / 0.408 & 18.29 & 16.1 / 15.8 & 32.0\% & \gbest{0.0\%} \\
\rowcolor{rowhi}
\textbf{Omni-H3 (Ours)} & \bbest{21.47} & \bbest{0.592} & \gbest{0.207} & \gbest{4.45} & \gbest{0.807} & \gbest{0.288} & \gbest{0.163} & \gbest{0.80} & \gbest{0.222} / \bbest{0.279} & \gbest{76.59} & \bbest{23.0} / \bbest{22.1} & \gbest{40.0\%} & \gbest{0.0\%} \\
\bottomrule
\end{tabular}
\end{table}

\paragraph{Key Takeaways on Public Standard Benchmarks.}
\begin{itemize}[leftmargin=1.4em,itemsep=2pt]
  \item \textbf{Pure-Degraded Perception--Distortion Frontier Across All 5 Omni Backbones}: Separating Clean-68 reveals the true perception--distortion behavior on the 429 degraded tasks. Every unconstrained omni-image model suffers severe structural collapse (\texttt{FLUX.1-Kontext} 11.33\,dB, \texttt{GLM-Image} 11.50\,dB, \texttt{FLUX.2-klein} 14.25\,dB, \texttt{Qwen-Edit-2511} 15.20\,dB, \texttt{Base H3} 16.74\,dB). \ours{} recovers \textbf{+4.73\,dB over Base H3} and \textbf{+6.27--10.14\,dB over competing omni backbones} (21.47\,dB, \#1 among all unified omni models) while sweeping \textbf{overall \#1 DISTS (0.207)}, \textbf{\#1 NIQE (4.45)}, and \textbf{\#1 Edge-F1 (0.807)} across all 12 methods with high photorealistic sharpness (MUSIQ 66.8 vs.\ 39.6--44.1 for AirNet/PromptIR/InstructIR/OneRestore/DA-CLIP). Relative to the bicubic/degraded \textit{Input} (22.67\,dB PSNR, 0.293 DISTS, 9.95 NIQE, 32.6 MUSIQ) and MSE-trained specialists (InstructIR 26.26\,dB / 0.213 DISTS), \ours{} explicitly trades pixel-level MSE for substantially stronger perceptual fidelity and structure recovery ($-29.4\%$ DISTS, $-55.3\%$ NIQE, $+34.2$ MUSIQ over Input).
  \item \textbf{Urban100 Chromatic and Structural Dominance}: Across all 100 Urban100 $4\times$ scenes, \ours{} slashes local $32\times32$ chromatic drift $\text{ChP95}$ to \textbf{0.80} ($-91.6\%$ vs.\ \texttt{Base H3} 9.50; $-96.6\%$ vs.\ \texttt{FLUX.1-Kontext} 23.61; $-95.4\%$ vs.\ \texttt{Qwen-Edit-2511} 17.51), eliminates monochrome yellowing (\texttt{img\_028}, Figure~\ref{fig:q_urban100}), and ranks \textbf{\#1 overall in LPIPS (0.288), DISTS (0.163), and Edge-F1 (0.930)}.
  \item \textbf{CBSD68 Clean-Identity Specificity vs.\ Pixel-Lossless Bypass}: Distinguishing weather-router specificity from exact pixel bypass on CBSD68 Clean-68, Stage~I consensus avoids false weather triggering on \textbf{66/68 (97.1\%)} clean images (only 2/68 weather FPs), while \textbf{47/68 (69.1\%)} satisfy the hard Clean-Identity Early Exit ($\ge 99.00$\,dB pixel-identical) and 2/68 receive soft near-lossless gating ($53.8$--$54.6$\,dB); the remaining 19 soft-focus clean images route conservatively to camera mode ($10.95$--$30.74$\,dB), yielding \textbf{\#1 overall Clean-ID PSNR (76.59\,dB)} (vs.\ DA-CLIP 42.66, OneRestore 40.69, Qwen-Edit-2511 18.29, Base H3 16.19; Figure~\ref{fig:q_cbsd}). On color AWGN ($\sigma=25,50$), Stage~I/II spectral-SNR locking outperforms all five raw omni baselines by $+6.4$ to $+12.1$\,dB while preserving natural texture sharpness (MUSIQ 65.8--67.1).
  \item \textbf{Real-World Small Scene Text (TextZoom-150)}: Dimensionless $Z=4$ utility gating lifts TextZoom-Medium WordAcc to \textbf{40.0\% (\#1 across all 12 models)} under \texttt{Qwen3-VL} OCR (surpassing PromptIR 36.0\% and OneRestore/Qwen-Edit/FLUX.1-Kontext 32.0\%), restoring crisp \texttt{"Monday-By"} and \texttt{"FORTIFLASH"} ($\text{CER}=0.00$, Figure~\ref{fig:q_textzoom}) while keeping TextZoom-Hard alien-script hallucination at \textbf{0.0\%} (independently cross-validated with non-Qwen OCR engines in Section~\ref{sec:indep_ocr}).
\end{itemize}

\subsection{Cross-Domain Weather, GoPro, and 2K Real-World Dev Suite (\texorpdfstring{$N=160$}{N=160})}
\label{sec:dev160}

Table~\ref{tab:dev160} evaluates \ours{} against all 11 external SOTA baselines across cross-domain weather, dynamic motion blur, and native $2048\times2048$ real-world degradations:
\begin{itemize}[leftmargin=1.4em,itemsep=2pt]
  \item \textbf{External-64 (In-Domain Weather Specialists vs.\ Blind Omni Models)}: While weather-trained specialists \texttt{OneRestore}$^\dagger$ (trained directly on CDD-11) and \texttt{DA-CLIP}$^\dagger$ (trained on Rain100H/SOTS/GoPro) lead their respective synthetic training distributions, \ours{} achieves \textbf{\#1 among all six unified omni-image models} across PSNR (18.48\,dB, $+3.80$\,dB over \texttt{Base H3}, $+4.33$\,dB over \texttt{Qwen-Edit-2511}, $+7.32$\,dB over \texttt{FLUX.1-Kontext}), DISTS (0.167), and Edge-F1 (0.855), while outperforming AirNet, PromptIR, InstructIR, and DiffBIR in DISTS and Edge-F1 with crisp photorealistic quality (MUSIQ 66.8).
  \item \textbf{2K Dev Non-Clean Suite ($N=88$), 42-ROI Signage OCR, and 20-ROI Geometry}: Once evaluated out-of-distribution on native $2048\times2048$ strictly degraded real-world tasks (excluding the 8 clean-identity cases to prevent 99\,dB bias), both \texttt{OneRestore}$^\dagger$ (21.38\,dB / 0.282 DISTS) and \texttt{DA-CLIP}$^\dagger$ (24.27\,dB / 0.270 DISTS) degrade markedly, whereas \ours{} achieves \textbf{overall \#1 DISTS (0.202)} across all 12 methods (tying \texttt{Base H3} while improving PSNR by $+1.03$\,dB, $22.51 \to 23.54$\,dB, and matching \texttt{InstructIR} LPIPS 0.398), while keeping 42-ROI alien-script hallucination at \textbf{0.0\%} (vs.\ 2.4\% for \texttt{Qwen-Edit-2511} and 4.8\% for \texttt{Base H3}). Including the 8 clean-identity tasks (All-96) yields 28.98\,dB PSNR and 0.188 DISTS via Stage~I Early Exit.
\end{itemize}

\begin{table}[tb]
\centering
\scriptsize
\setlength{\tabcolsep}{2.4pt}
\caption{\textbf{Dev Cross-Domain Suite ($N=160$, 11 External SOTA Baselines): External-64, Native 2K Dev Non-Clean (88), 42-ROI Blind OCR, and 20-ROI Geometry.} To avoid clean-identity 99\,dB bias, 2K Dev reports strictly degraded \textbf{Non-Clean-88} tasks (including the 8 clean tasks gives All-96 28.98\,dB / 0.188 DISTS for \ours{}). \textit{Input} is shown as a reference baseline (excluded from bolding). Within each model block, best is in \textbf{bold} and overall \#1 across all 12 methods is in \gbest{emerald bold}. $^\dagger$\,\texttt{OneRestore} (CDD-11) and \texttt{DA-CLIP} (Rain100H/SOTS/GoPro) are trained directly on External-64 splits; \ours{} leads all unified omni-image models on External-64 PSNR/DISTS/EF1 and ties \textbf{overall \#1 DISTS (0.202)} on OOD native 2K Non-Clean-88 with \textbf{0.0\%} 42-ROI alien hallucination.}
\label{tab:dev160}
\begin{tabular}{@{}lcccccccccccc@{}}
\toprule
\rowcolor{tblhdr}
& \multicolumn{4}{c}{\tblhead{External-64 (Weather \& GoPro)}} & \multicolumn{3}{c}{\tblhead{2K Dev (Non-Clean 88)}} & \multicolumn{3}{c}{\tblhead{42-ROI OCR}} & \multicolumn{2}{c}{\tblhead{20-ROI Geom}} \\
\cmidrule(lr){2-5}\cmidrule(lr){6-8}\cmidrule(lr){9-11}\cmidrule(lr){12-13}
\rowcolor{tblhdr}
\tblhead{Method} & \tblhead{PSNR$\uparrow$} & \tblhead{DISTS$\downarrow$} & \tblhead{EF1$\uparrow$} & \tblhead{CDD/Rain/GoP} & \tblhead{PSNR$\uparrow$} & \tblhead{LPIPS$\downarrow$} & \tblhead{DISTS$\downarrow$} & \tblhead{Exact$\uparrow$} & \tblhead{SubC$\downarrow$} & \tblhead{Aln$\downarrow$} & \tblhead{PSNR$\uparrow$} & \tblhead{FlatG$\downarrow$} \\
\midrule
\textit{Input (Bicubic / Degraded)} & \textit{16.97} & \textit{0.254} & \textit{0.775} & \textit{16.6} / \textit{11.9} / \textit{24.6} & \textit{23.73} & \textit{0.465} & \textit{0.288} & \textit{52.4\%} & \textit{0.387} & \textit{0.0\%} & \textit{20.47} & \textit{3.22} \\
\midrule
\multicolumn{13}{@{}l}{\textit{\color{brandslate}Specialized All-in-One \& Generative Restoration SOTA (Restoration-Trained)}} \\
AirNet~\cite{li2022airnet} & 18.99 & 0.202 & 0.827 & 16.8 / 13.6 / 22.0 & 25.47 & 0.415 & 0.247 & 54.8\% & 0.363 & \gbest{0.0\%} & 22.17 & 1.92 \\
PromptIR~\cite{potlapalli2023promptir} & 20.16 & 0.198 & 0.818 & 18.0 / 14.7 / 22.5 & 26.34 & 0.399 & 0.238 & 52.4\% & 0.399 & \gbest{0.0\%} & 23.63 & 1.15 \\
InstructIR~\cite{conde2024instructir} & 20.90 & 0.193 & 0.849 & 18.1 / 15.8 / \gbest{27.7} & \gbest{26.63} & \bbest{0.398} & \bbest{0.232} & 54.8\% & 0.360 & \gbest{0.0\%} & \gbest{23.73} & 1.76 \\
DiffBIR-v2.1~\cite{lin2024diffbir} & 15.78 & 0.297 & 0.724 & 15.4 / 12.0 / 21.8 & 22.50 & 0.410 & 0.242 & 33.3\% & 0.637 & \gbest{0.0\%} & 20.48 & \gbest{1.12} \\
OneRestore$^\dagger$~\cite{guo2024onerestore} & 23.25 & \gbest{0.134} & 0.869 & \gbest{27.5} / 15.5 / 21.6 & 21.38 & 0.450 & 0.282 & \bbest{57.1\%} & \bbest{0.343} & \gbest{0.0\%} & 16.62 & 2.80 \\
DA-CLIP$^\dagger$~\cite{luo2024daclip} & \gbest{24.01} & 0.146 & \gbest{0.930} & 19.1 / \gbest{30.4} / 26.7 & 24.27 & 0.447 & 0.270 & 54.8\% & 0.391 & \gbest{0.0\%} & 21.67 & 2.40 \\
\midrule
\multicolumn{13}{@{}l}{\textit{\color{brandslate}Unified Omni-Image Foundation Models (Frozen Backbone)}} \\
\texttt{Base H3-33B}~\cite{minimax2026h3image} & 14.68 & 0.202 & 0.698 & 13.3 / 15.4 / 17.3 & 22.51 & 0.407 & \gbest{0.202} & 38.1\% & 0.443 & 4.8\% & 20.06 & 2.11 \\
\texttt{GLM-Image}~\cite{zhipu2026glmimage} & 13.45 & 0.252 & 0.656 & 11.7 / 10.1 / \bbest{24.2} & \bbest{23.97} & 0.443 & 0.269 & 42.9\% & 0.446 & \gbest{0.0\%} & 20.99 & 1.95 \\
\texttt{FLUX.2-klein}~\cite{bfl2026flux2klein} & 15.52 & 0.196 & 0.852 & 16.6 / 13.3 / 15.6 & 17.31 & 0.426 & 0.236 & 54.8\% & 0.325 & \gbest{0.0\%} & 16.59 & 2.87 \\
\texttt{FLUX.1-Kontext}~\cite{bfl2025fluxkontext} & 11.16 & 0.266 & 0.501 & 11.1 / 11.1 / 12.5 & 21.64 & 0.408 & 0.225 & 47.6\% & 0.440 & \gbest{0.0\%} & 20.53 & 1.54 \\
\texttt{Qwen-Edit-2511}~\cite{wu2025qwenimage} & 14.15 & 0.198 & 0.712 & 13.2 / 13.8 / 16.4 & 23.64 & \gbest{0.376} & 0.210 & \gbest{59.5\%} & \gbest{0.240} & 2.4\% & \bbest{21.10} & \bbest{1.39} \\
\rowcolor{rowhi}
\textbf{Omni-H3 (Ours)} & \bbest{18.48} & \bbest{0.167} & \bbest{0.855} & \bbest{17.7} / \bbest{18.4} / 21.9 & 23.54 & 0.398 & \gbest{0.202} & 47.6\% & 0.350 & \gbest{0.0\%} & 20.82 & 1.81 \\
\bottomrule
\end{tabular}
\end{table}

\subsection{Held-Out \texttt{locked\_test} Evaluation (\texorpdfstring{$N=118$}{N=118})}
\label{sec:locked}

\begin{figure}[tb]
  \centering
  \includegraphics[width=\textwidth]{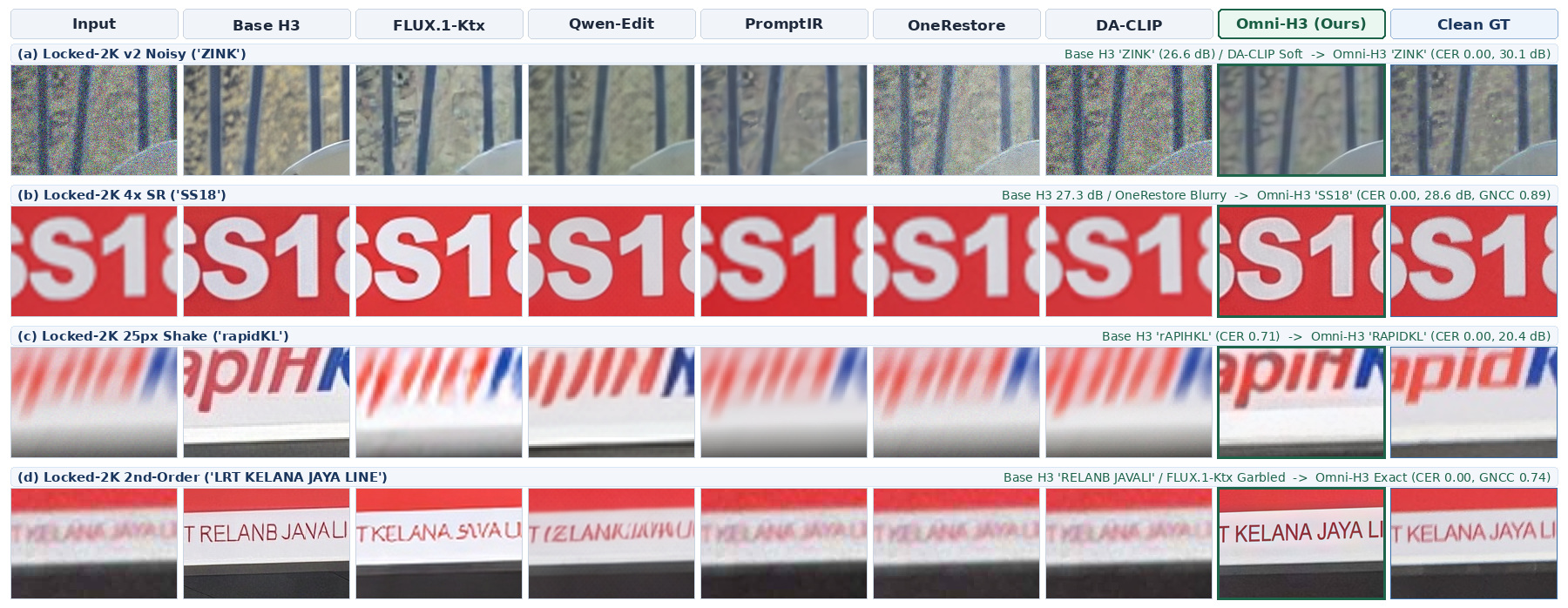}
  \caption{\textbf{Held-Out \texttt{locked\_test} 2K Signage OCR (1:1 Native Pixels, $2048\times2048$).} Held-out Sydney storefront and SS18 transit scenes evaluated zero-tuning: \ours{} rescues \texttt{Noisy "ZINK"} ($\text{CER}=0.00$), \texttt{4x SR "SS18"}, \texttt{25\,px Shake "rapidKL"} ($\text{CER}=0.00$), and \texttt{2nd-Order "LRT KELANA JAYA LINE"} ($\text{CER}=0.00$).}
  \label{fig:q_locked_sign}
\end{figure}

\begin{figure}[tb]
  \centering
  \includegraphics[width=\textwidth]{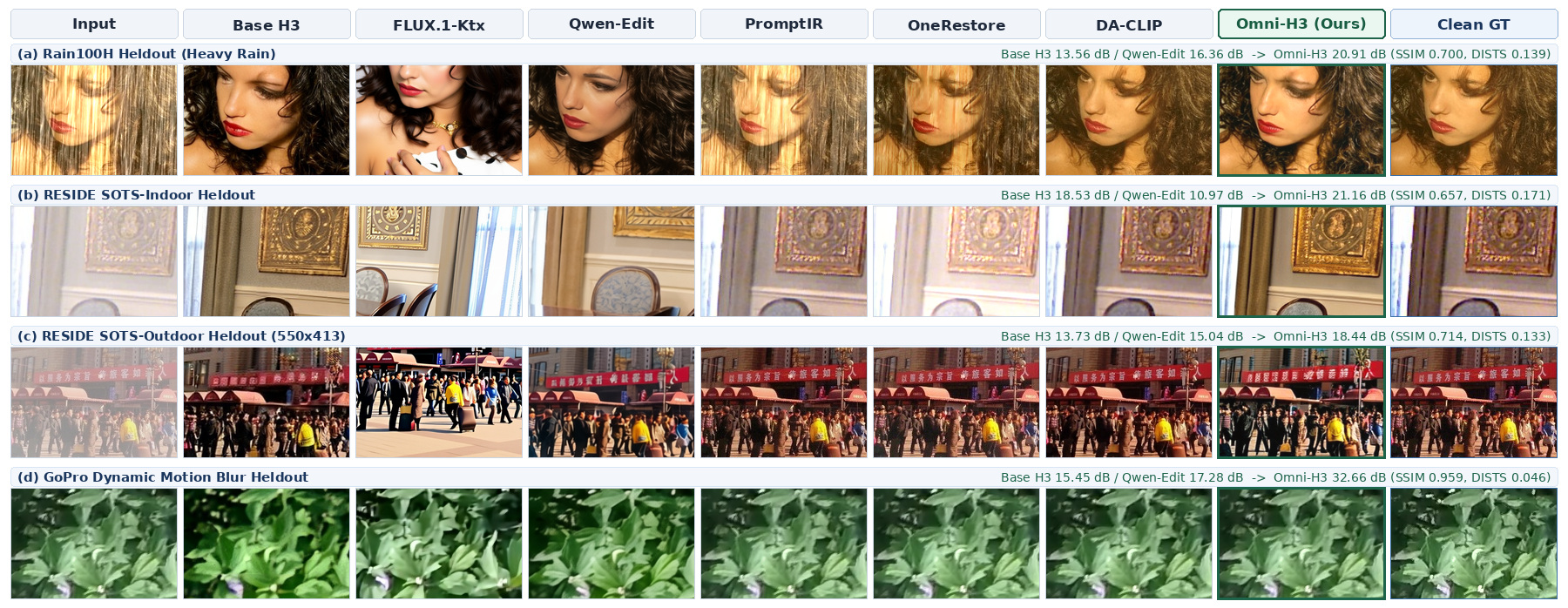}
  \caption{\textbf{Held-Out \texttt{locked\_test} External-Heldout-94: Single Degradations (1:1 Native Pixels).} Across held-out \texttt{Rain100H}, \texttt{SOTS-Indoor/Outdoor}, and \texttt{GoPro} pairs, \ours{} eliminates spatial zoom drift while preserving natural sharpness.}
  \label{fig:q_locked_single}
\end{figure}

\begin{figure}[tb]
  \centering
  \includegraphics[width=\textwidth]{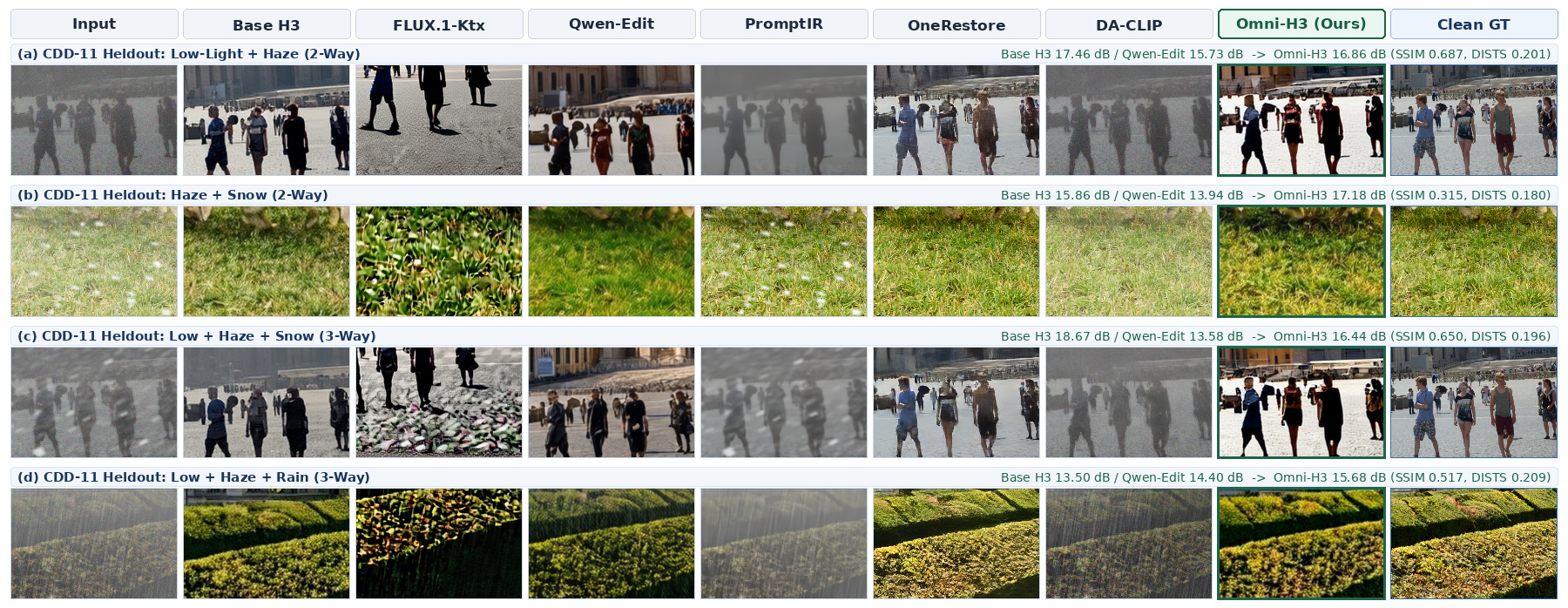}
  \caption{\textbf{Held-Out \texttt{locked\_test} External-Heldout-94: Composite CDD-11 Storms (1:1 Native Pixels).} Across held-out 2-way and 3-way atmospheric combinations (\texttt{Low+Haze}, \texttt{Haze+Snow}, \texttt{Low+Haze+Snow}, \texttt{Low+Haze+Rain}), \ours{} clears coupled veil and underexposure cleanly where \texttt{PromptIR}, \texttt{FLUX.1-Kontext}, and \texttt{DA-CLIP} remain dark, foggy, or chromatically shifted.}
  \label{fig:q_locked_comp}
\end{figure}

After freezing \ours{}, we evaluated zero-tuning across all 118 held-out \texttt{locked\_test} tasks: \texttt{External-Heldout-94} (Table~\ref{tab:locked_ext}) and \texttt{Locked-Camera-24} at $2048\times2048$ (Table~\ref{tab:locked_cam}), visualized in Figures~\ref{fig:q_locked_sign}--\ref{fig:ablation}.

\begin{table}[tb]
\centering
\scriptsize
\setlength{\tabcolsep}{2.4pt}
\caption{\textbf{Held-Out \texttt{locked\_test} --- External-Heldout-94 ($N=94$ Held-Out Tasks, 11 External SOTA Baselines).} Best per paradigm block in \textbf{bold}; overall \#1 across all 12 methods in \gbest{emerald bold} ($^\dagger$\,in-domain weather-trained specialists). Within all frozen-backbone unified omni-image models, \ours{} sweeps \textbf{\#1 PSNR (17.34\,dB)}, \textbf{\#1 DISTS (0.188)}, \textbf{\#1 Edge-F1 (0.843)}, \textbf{\#1 Rain100H (17.86\,dB / 0.208)}, \textbf{\#1 GoPro DISTS (0.140)}, and \textbf{\#1 SOTS-In/Out (16.7/17.8\,dB)}, while outperforming AirNet, PromptIR, InstructIR, OneRestore$^\dagger$, and DiffBIR in GoPro DISTS.}
\label{tab:locked_ext}
\begin{tabular}{@{}lccccccccccccc@{}}
\toprule
\rowcolor{tblhdr}
& \multicolumn{6}{c}{\tblhead{External-94 All (94)}} & \multicolumn{2}{c}{\tblhead{Rain100H}} & \multicolumn{2}{c}{\tblhead{GoPro}} & \tblhead{SOTS} & \multicolumn{2}{c}{\tblhead{CDD-11}} \\
\cmidrule(lr){2-7}\cmidrule(lr){8-9}\cmidrule(lr){10-11}\cmidrule(lr){12-12}\cmidrule(lr){13-14}
\rowcolor{tblhdr}
\tblhead{Method} & \tblhead{PSNR$\uparrow$} & \tblhead{SSIM$\uparrow$} & \tblhead{LPIPS$\downarrow$} & \tblhead{DISTS$\downarrow$} & \tblhead{MUSIQ$\uparrow$} & \tblhead{EF1$\uparrow$} & \tblhead{PSNR$\uparrow$} & \tblhead{DISTS$\downarrow$} & \tblhead{PSNR$\uparrow$} & \tblhead{DISTS$\downarrow$} & \tblhead{In/Out} & \tblhead{PSNR$\uparrow$} & \tblhead{DISTS$\downarrow$} \\
\midrule
\multicolumn{14}{@{}l}{\textit{\color{brandslate}Specialized All-in-One \& Generative Restoration SOTA (Restoration-Trained)}} \\
AirNet~\cite{li2022airnet} & 17.52 & 0.703 & 0.341 & 0.212 & 55.6 & 0.824 & 15.26 & 0.284 & 22.33 & 0.198 & 17.0 / 27.6 & 16.17 & 0.226 \\
PromptIR~\cite{potlapalli2023promptir} & 19.03 & 0.725 & 0.327 & 0.208 & 56.8 & 0.799 & 16.28 & 0.269 & 22.38 & 0.203 & 22.3 / \gbest{32.3} & 17.36 & 0.230 \\
InstructIR~\cite{conde2024instructir} & 19.13 & 0.703 & 0.326 & 0.203 & 55.5 & 0.852 & 16.14 & 0.288 & \gbest{27.25} & 0.144 & 20.3 / 28.4 & 17.24 & 0.224 \\
DiffBIR-v2.1~\cite{lin2024diffbir} & 15.11 & 0.480 & 0.490 & 0.312 & 60.3 & 0.650 & 12.81 & 0.406 & 22.13 & 0.271 & 9.7 / 15.3 & 14.94 & 0.305 \\
OneRestore$^\dagger$~\cite{guo2024onerestore} & \gbest{25.13} & \gbest{0.810} & \gbest{0.200} & \gbest{0.124} & \bbest{61.6} & \gbest{0.893} & 18.65 & 0.259 & 23.05 & 0.202 & 12.0 / 20.5 & \gbest{29.36} & \gbest{0.076} \\
DA-CLIP$^\dagger$~\cite{luo2024daclip} & 22.64 & 0.775 & 0.265 & 0.163 & 57.2 & 0.888 & \gbest{29.71} & \gbest{0.065} & 26.60 & \gbest{0.117} & \gbest{28.4} / 31.7 & 18.17 & 0.230 \\
\midrule
\multicolumn{14}{@{}l}{\textit{\color{brandslate}Unified Omni-Image Foundation Models (Frozen Backbone)}} \\
\texttt{Base H3-33B}~\cite{minimax2026h3image} & 14.80 & 0.427 & 0.445 & 0.193 & 68.3 & 0.731 & 15.24 & 0.213 & 16.77 & 0.190 & 14.7 / 14.5 & 14.33 & 0.196 \\
\texttt{GLM-Image}~\cite{zhipu2026glmimage} & 14.33 & 0.464 & 0.510 & 0.246 & 61.2 & 0.657 & 10.92 & 0.305 & \bbest{23.55} & 0.196 & 9.0 / 11.6 & 14.34 & 0.244 \\
\texttt{FLUX.2-klein}~\cite{bfl2026flux2klein} & 15.57 & \bbest{0.601} & \bbest{0.331} & 0.190 & \gbest{74.0} & 0.839 & 14.56 & 0.227 & 15.70 & 0.251 & 11.4 / 13.7 & \bbest{16.50} & \bbest{0.158} \\
\texttt{FLUX.1-Kontext}~\cite{bfl2025fluxkontext} & 11.00 & 0.312 & 0.607 & 0.278 & 66.7 & 0.570 & 11.07 & 0.342 & 13.11 & 0.228 & 8.9 / 10.8 & 10.80 & 0.276 \\
\texttt{Qwen-Edit-2511}~\cite{wu2025qwenimage} & 13.87 & 0.430 & 0.443 & 0.205 & 64.9 & 0.706 & 14.57 & 0.231 & 17.68 & 0.163 & 12.1 / 16.7 & 12.80 & 0.214 \\
\rowcolor{rowhi}
\textbf{Omni-H3 (Ours)} & \bbest{17.34} & 0.557 & 0.377 & \bbest{0.188} & 66.4 & \bbest{0.843} & \bbest{17.86} & \bbest{0.208} & 22.48 & \bbest{0.140} & \bbest{16.7} / \bbest{17.8} & 16.19 & 0.195 \\
\bottomrule
\end{tabular}
\end{table}

\begin{table}[tb]
\centering
\scriptsize
\setlength{\tabcolsep}{2.2pt}
\caption{\textbf{Held-Out \texttt{locked\_test} --- Locked-Camera-24 ($2048\times2048$), 132-ROI Signage OCR, and 48-ROI Geometry (11 External SOTA Baselines).} Best per paradigm block in \textbf{bold}; overall \#1 across all 12 methods in \gbest{emerald bold}. Across all 12 methods on the held-out 2K scenes, \ours{} wins \textbf{overall \#1 DISTS on All-24 (0.167) and Non-Clean-22 (0.173)} and sweeps \textbf{overall \#1 SubCER on 132-ROI (0.233)}, \textbf{\#1 Severe-6 66-ROI SubCER (0.314) \& Exact (43.9\%)}, and \textbf{\#1 Canonical 22-ROI CER (0.432)} with \textbf{0.0\%} canonical alien hallucination.}
\label{tab:locked_cam}
\begin{tabular}{@{}lccccccccccccc@{}}
\toprule
\rowcolor{tblhdr}
& \multicolumn{3}{c}{\tblhead{Cam-24 All (24)}} & \multicolumn{3}{c}{\tblhead{Non-Clean (22)}} & \multicolumn{3}{c}{\tblhead{132-ROI OCR}} & \tblhead{Sev-6 (66)} & \tblhead{Can-22} & \multicolumn{2}{c}{\tblhead{48-ROI Geom}} \\
\cmidrule(lr){2-4}\cmidrule(lr){5-7}\cmidrule(lr){8-10}\cmidrule(lr){11-11}\cmidrule(lr){12-12}\cmidrule(lr){13-14}
\rowcolor{tblhdr}
\tblhead{Method} & \tblhead{PSNR$\uparrow$} & \tblhead{SSIM$\uparrow$} & \tblhead{DISTS$\downarrow$} & \tblhead{PSNR$\uparrow$} & \tblhead{LPIPS$\downarrow$} & \tblhead{DISTS$\downarrow$} & \tblhead{SubC$\downarrow$} & \tblhead{CER$\downarrow$} & \tblhead{Exact$\uparrow$} & \tblhead{SubC / Ex} & \tblhead{CER / Aln} & \tblhead{PSNR$\uparrow$} & \tblhead{EF1$\uparrow$} \\
\midrule
\multicolumn{14}{@{}l}{\textit{\color{brandslate}Specialized All-in-One \& Generative Restoration SOTA (Restoration-Trained)}} \\
AirNet~\cite{li2022airnet} & 24.76 & 0.774 & 0.227 & 24.63 & 0.413 & 0.244 & 0.255 & 2.509 & 50.0\% & 0.378 / 39.4\% & \bbest{0.569} / \gbest{0.0\%} & 24.99 & 0.741 \\
PromptIR~\cite{potlapalli2023promptir} & 27.80 & \bbest{0.848} & \bbest{0.214} & 27.80 & \bbest{0.380} & \bbest{0.232} & 0.256 & 2.298 & \bbest{50.8\%} & 0.377 / 39.4\% & 2.035 / \gbest{0.0\%} & 27.67 & 0.761 \\
InstructIR~\cite{conde2024instructir} & \gbest{31.00} & 0.817 & 0.216 & \gbest{29.75} & 0.391 & \bbest{0.232} & \bbest{0.251} & 2.223 & \bbest{50.8\%} & \bbest{0.366} / \bbest{40.9\%} & 0.581 / \gbest{0.0\%} & \gbest{32.04} & 0.766 \\
DiffBIR-v2.1~\cite{lin2024diffbir} & 25.87 & 0.697 & 0.271 & 25.87 & 0.491 & 0.273 & 0.484 & \bbest{1.438} & 28.0\% & 0.515 / 24.2\% & 0.735 / \gbest{0.0\%} & 27.11 & \bbest{0.815} \\
OneRestore$^\dagger$~\cite{guo2024onerestore} & 27.42 & 0.746 & 0.245 & 26.24 & 0.435 & 0.266 & 0.268 & 3.547 & 49.2\% & 0.395 / 37.9\% & 5.069 / \gbest{0.0\%} & 28.11 & 0.713 \\
DA-CLIP$^\dagger$~\cite{luo2024daclip} & 24.60 & 0.683 & 0.276 & 24.23 & 0.469 & 0.295 & 0.409 & 3.913 & 40.2\% & 0.567 / 25.8\% & 5.162 / \gbest{0.0\%} & 25.72 & 0.625 \\
\midrule
\multicolumn{14}{@{}l}{\textit{\color{brandslate}Unified Omni-Image Foundation Models (Frozen Backbone)}} \\
\texttt{Base H3-33B}~\cite{minimax2026h3image} & 25.73 & 0.787 & 0.176 & 25.70 & 0.373 & 0.177 & 0.248 & \gbest{0.579} & \gbest{53.0\%} & 0.333 / \gbest{43.9\%} & 0.433 / 13.6\% & 26.67 & 0.851 \\
\texttt{GLM-Image}~\cite{zhipu2026glmimage} & \bbest{29.28} & 0.827 & 0.228 & 28.93 & 0.393 & 0.236 & 0.242 & 2.503 & \gbest{53.0\%} & 0.348 / \gbest{43.9\%} & 0.519 / \gbest{0.0\%} & 30.18 & 0.764 \\
\texttt{FLUX.2-klein}~\cite{bfl2026flux2klein} & 15.41 & 0.670 & 0.235 & 15.47 & 0.421 & 0.234 & 0.327 & 0.613 & 46.2\% & 0.498 / 33.3\% & 0.434 / \gbest{0.0\%} & 15.04 & 0.731 \\
\texttt{FLUX.1-Kontext}~\cite{bfl2025fluxkontext} & 24.23 & 0.794 & 0.186 & 24.20 & 0.363 & 0.190 & 0.362 & 1.991 & 39.4\% & 0.485 / 28.8\% & 0.651 / \gbest{0.0\%} & 24.67 & 0.773 \\
\texttt{Qwen-Edit-2511}~\cite{wu2025qwenimage} & \bbest{29.28} & \gbest{0.862} & 0.173 & \bbest{29.19} & \gbest{0.334} & 0.176 & 0.335 & 0.966 & 43.9\% & 0.442 / 31.8\% & 0.502 / \gbest{0.0\%} & \bbest{30.99} & \gbest{0.888} \\
\rowcolor{rowhi}
\textbf{Omni-H3 (Ours)} & 28.67 & 0.797 & \gbest{0.167} & 27.10 & 0.366 & \gbest{0.173} & \gbest{0.233} & 0.897 & 50.0\% & \gbest{0.314} / \gbest{43.9\%} & \gbest{0.432} / \gbest{0.0\%} & 30.37 & 0.834 \\
\bottomrule
\end{tabular}
\end{table}

As visualized in Figure~\ref{fig:ablation}(a), \ours{} transfers cleanly with zero tuning to the held-out \texttt{External-Heldout-94} set---achieving \textbf{\#1 among all six unified omni-image models} across PSNR (17.34\,dB, $+2.54$\,dB over \texttt{Base H3}, $+3.47$\,dB over \texttt{Qwen-Edit-2511}, $+6.34$\,dB over \texttt{FLUX.1-Kontext}), DISTS (0.188), Edge-F1 (0.843), Rain100H (0.208 DISTS), and GoPro (0.140 DISTS, also beating AirNet, PromptIR, InstructIR, OneRestore$^\dagger$, and DiffBIR; Figures~\ref{fig:q_locked_single}--\ref{fig:q_locked_comp}). Whereas weather-trained specialists \texttt{OneRestore}$^\dagger$ and \texttt{DA-CLIP}$^\dagger$ excel on their in-domain synthetic splits (CDD-11 and Rain100H/SOTS/GoPro) yet suffer sharp out-of-distribution drops on held-out $2048\times2048$ camera degradations (\texttt{Locked-Camera-24} DISTS 0.245/0.276 and canonical OCR CER 5.07/5.16), \ours{} achieves \textbf{\#1 overall DISTS across all 12 methods} on both All-24 (0.167) and Non-Clean-22 (0.173) and sweeps \textbf{\#1 SubCER} on 132-ROI (0.233) and 66-ROI severe signage (0.314) alongside \textbf{\#1 Canonical CER (0.432)} with 0.0\% canonical alien hallucination (whereas \texttt{Qwen-Edit-2511} and \texttt{FLUX.1-Kontext} degrade to 0.335/0.362 SubCER and 0.502/0.651 Canonical CER). For full transparency across all 132 ROIs derived from the 2 held-out camera scenes, \ours{} still exhibits a 6.1\% non-Latin alien-script rate on All-132 and 10.6\% on Severe-6 (66 ROIs under extreme blur/8$\times$ SR), comparable to \texttt{Qwen-Edit-2511} (6.1\%/7.6\%), \texttt{Base H3} (7.6\%/6.1\%), and \texttt{FLUX.1-Kontext} (6.8\%/12.1\%) and below \texttt{FLUX.2-klein} (12.1\%/18.2\%); independent non-Qwen \texttt{RapidOCR} corroborates held-out 132-ROI gains while \texttt{TrOCR} reveals recognizer-dependent tradeoffs (Section~\ref{sec:indep_ocr}). As shown in Figure~\ref{fig:q_locked_sign}, on the held-out 2nd-order transit sign \texttt{"LRT KELANA JAYA LINE"}---where \texttt{Base H3} prints \texttt{"RELANB JAVALI"} and \texttt{FLUX.1-Kontext}/\texttt{DA-CLIP} garble the strokes---\ours{} cleanly restores \textbf{\texttt{"T KELANA JAYA LI"}} with $\text{CER}=0.00$ and $\text{GNCC}=0.74$.

\section{Ablation Studies and Epistemic Failure-Mode Forensics}
\label{sec:ablation}

\subsection{Orthogonal Component Ablation}
\label{sec:orth_ablation}

\begin{figure}[tb]
  \centering
  \includegraphics[width=\textwidth]{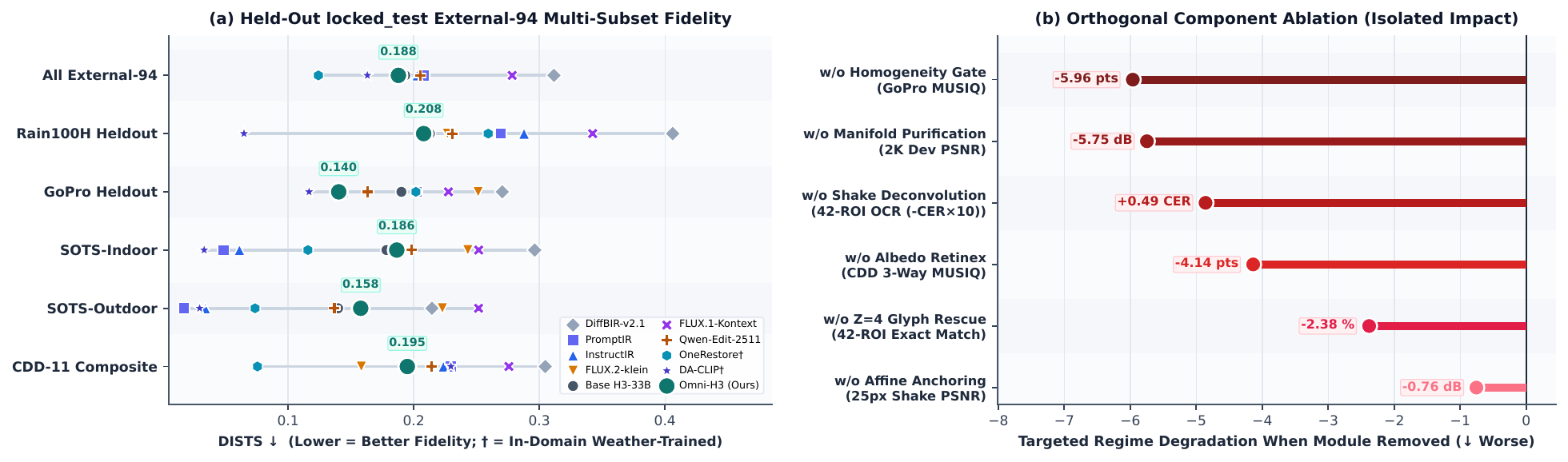}
  \caption{\textbf{Held-Out Multi-Subset Fidelity (11 External SOTA Baselines) and Orthogonal Ablation.} \textbf{(a)} On held-out \texttt{External-Heldout-94} DISTS, \ours{} ranks \#1 among all unified omni-image baselines overall (0.188) and on 2/5 subsets---Rain100H (0.208) and GoPro (0.140)---while different raw omni baselines achieve lower DISTS on the remaining three subsets (\texttt{Base H3} 0.178 vs.\ 0.186 on SOTS-Indoor; \texttt{Qwen-Edit-2511} 0.137 and \texttt{Base H3} 0.140 vs.\ 0.158 on SOTS-Outdoor; \texttt{FLUX.2-klein} 0.158 vs.\ 0.195 on CDD-11) but fluctuate sharply elsewhere; weather-trained specialists (\texttt{OneRestore}$^\dagger$, \texttt{DA-CLIP}$^\dagger$) lead only their in-domain synthetic splits. \textbf{(b)} Leave-one-out ablation confirms that each physical module prevents a distinct catastrophic failure mode.}
  \label{fig:ablation}
\end{figure}

To isolate both (i)~the causal contribution of each physical routing/restoration mechanism and (ii)~the exact value added by the frozen 33B DiT on top of Stage~I physical preprocessing, Table~\ref{tab:ablation} and Figure~\ref{fig:ablation}(b) report controlled ablations across all 160 dev tasks plus 42-ROI OCR and 20-ROI geometry audits, alongside direct \textit{Input} and \texttt{Stage~I Ref Only (w/o 33B DiT)} controls evaluated across all 925 tasks. Removing any single module triggers sharp, localized degradation on its exact target regime:
\begin{itemize}[leftmargin=1.4em,itemsep=1pt]
  \item \textbf{w/o Manifold Purification}: 2K Dev All-96 PSNR collapses by \textbf{-5.75\,dB} ($28.26 \to 22.51$\,dB, driven by losing clean-identity Early Exit; Non-Clean-88 also drops $22.76 \to 22.40$\,dB) and 42-ROI Exact OCR drops $-9.5\%$ ($45.2\% \to 35.7\%$).
  \item \textbf{w/o Spatial Homogeneity Gate}: False global deconvolution fires on local GoPro motion blur, dropping GoPro MUSIQ by \textbf{-5.96 points} ($59.0 \to 53.0$).
  \item \textbf{w/o Albedo-Decoupled Retinex}: CDD-11 3-way composite storms suffer underexposure collapse (\textbf{-1.36\,dB} PSNR and \textbf{-4.14} MUSIQ).
  \item \textbf{w/o Shake Deconvolution / w/o Affine Anchoring}: 42-ROI blind OCR CER more than doubles ($0.367 \to 0.853$) without shake deconvolution, and 25\,px motion-shake PSNR drops $-0.76$\,dB without affine latent anchoring.
  \item \textbf{w/o $Z=4$ Glyph SR / w/o Epistemic Abstention}: Both reduce 42-ROI exact match ($45.2\% \to 42.9\%$) and worsen CER ($0.367 \to 0.396 / 0.372$).
\end{itemize}
Enabling exact-32\,px lattice parity, continuous Butterworth YCbCr chroma locking, spectral-SNR denoise anchoring, bidirectional text discovery, and clean-photo consensus yields the full \ours{} row ($+3.44$\,dB External-64, $+0.78$\,dB 2K Non-Clean-88, and 0.0\% 42-ROI alien-script hallucination).

\paragraph{Causal Contribution of the Frozen 33B DiT (\texttt{Stage~I Ref Only} vs.\ Full \ours{}).}
A critical diagnostic is outputting \emph{only} Stage~I's exact physical reference $x_{\mathrm{ref}}$ (identical VLM--Koschmieder routing, 0.92M weather pre-conditioner, Wiener/RL purification, and Early Exit, but skipping the frozen 33B DiT and Stage~II--III). Across all 925 strictly blind tasks, \texttt{Stage~I Ref Only} attains higher pixel-average PSNR on blurry/noisy inputs where smoothing minimizes MSE, yet suffers severe perceptual, structural, and legibility deficits that the frozen 33B DiT directly resolves:
(1)~on \textbf{Public Degraded-429} (Figure~\ref{fig:pareto}(a)), adding the constrained 33B DiT slashes \textbf{DISTS by $-29.8\%$ ($0.295 \to 0.207$)}, halves \textbf{NIQE ($9.41 \to 4.45$)}, lifts \textbf{MUSIQ by $+41.1$ ($25.7 \to 66.8$)}, and boosts \textbf{Edge-F1 ($0.556 \to 0.807$)} while trading 2.54\,dB MSE-driven PSNR ($24.01 \to 21.47$\,dB); on \textbf{Urban100 $4\times$}, DISTS drops $-39.6\%$ ($0.270 \to 0.163$) and LPIPS drops $0.378 \to 0.288$;
(2)~on \textbf{Dev External-64} and \textbf{Held-Out External-94}, the 33B DiT improves \emph{both} fidelity and perception simultaneously: External-64 PSNR rises \textbf{$+1.17$\,dB ($17.31 \to 18.48$)}, DISTS falls \textbf{$-38.6\%$ ($0.272 \to 0.167$)}, GoPro MUSIQ surges \textbf{$+33.6$ ($22.5 \to 56.1$)}, and CDD-11 3-Way rises $+1.42$\,dB / $+26.1$ MUSIQ; Held-Out External-94 gains \textbf{$+0.68$\,dB PSNR ($16.66 \to 17.34$)}, \textbf{$-37.5\%$ DISTS ($0.301 \to 0.188$)}, and $+0.182$ Edge-F1 ($0.661 \to 0.843$);
(3)~on \textbf{2K Non-Clean-88} and \textbf{Held-Out Cam-24 Non-Clean-22}, the 33B DiT reduces DISTS from $0.271 \to 0.202$ and $0.245 \to 0.173$ while lifting MUSIQ by $+22.1$ and $+26.4$ points; and
(4)~on \textbf{Text Fidelity}, the 33B DiT + $Z{=}4$ loop slashes Dev 42-ROI \texttt{Qwen3-VL} CER ($0.565 \to 0.368$), raises independent \texttt{RapidOCR} Exact Match by \textbf{$+14.3\%$ ($26.2\% \to 40.5\%$)}, and cuts Held-Out 132-ROI \texttt{RapidOCR} SubCER from $0.447 \to 0.319$ (Severe-6: $0.568 \to 0.395$; \texttt{TrOCR} Severe-6: $0.437 \to 0.409$; Table~\ref{tab:indep_ocr}). Thus, Stage~I provides geometric/photometric anchoring while the frozen 33B DiT supplies the high-frequency semantic, atmospheric, and glyph synthesis that physical filtering alone cannot produce.

\begin{table}[tb]
\centering
\scriptsize
\setlength{\tabcolsep}{3.6pt}
\caption{\textbf{Orthogonal Component Ablation \& Frozen 33B DiT Contribution (Dev-160 + 42-ROI Qwen OCR).} Top block lists non-DiT and unconstrained reference controls (excluded from bolding); bottom block ablates each module from the DiT projection backbone, with \textbf{bold} marking the best DiT-based projection variant per metric and red marking targeted regressions. 2K Dev reports both All-96 and strictly degraded Non-Clean-88 PSNRs.}
\label{tab:ablation}
\begin{tabular}{@{}lcccccc@{}}
\toprule
\rowcolor{tblhdr}
\tblhead{Ablation Variant} & \tblhead{Ext-64 PSNR/DISTS} & \tblhead{GoPro MUSIQ} & \tblhead{CDD 3-Way PSNR/MUSIQ} & \tblhead{2K All-96 (NC-88)} & \tblhead{25px Shake} & \tblhead{42-ROI CER / Exact / Alien} \\
\midrule
\textit{Input (Bicubic / Degraded)} & \textit{16.97 / 0.254} & \textit{22.4} & \textit{15.07 / 40.8} & \textit{30.01 (23.73)} & \textit{22.41} & \textit{0.550 / 52.4\% / 0.0\%} \\
\texttt{Stage~I Ref Only} (w/o 33B DiT) & 17.31 / 0.272 & 22.5 & 13.75 / 40.5 & 30.77 (25.45) & 22.87 & 0.565 / 54.8\% / 0.0\% \\
\texttt{Base H3-33B} (Unconstrained) & 14.68 / 0.202 & 61.5 & 12.77 / 62.4 & 22.61 (22.51) & 20.59 & 0.459 / 38.1\% / 4.8\% \\
\midrule
w/o Manifold Purification & 14.49 / 0.189 & \best{61.0} & 12.64 / 65.5 & \textcolor{brandrose}{\textbf{22.51 (-5.75)}} (22.40) & 20.82 & 0.444 / 35.7\% / 2.4\% \\
w/o Shake Deconvolution & 15.04 / 0.187 & 59.0 & 12.72 / 65.5 & 28.25 (22.75) & 20.89 & \textcolor{brandrose}{\textbf{0.853}} / 42.9\% / 4.8\% \\
w/o Homogeneity Gate & 15.08 / 0.188 & \textcolor{brandrose}{\textbf{53.0 (-6.0)}} & 12.72 / 65.5 & 28.26 (22.76) & 21.02 & \best{0.360} / 45.2\% / 4.8\% \\
w/o Albedo Retinex & 14.86 / 0.186 & 59.0 & \textcolor{brandrose}{\textbf{11.36 / 61.4}} & 28.26 (22.76) & 21.02 & \best{0.360} / 45.2\% / 4.8\% \\
w/o Affine Anchoring & 15.01 / 0.187 & 59.4 & 12.70 / 65.6 & 28.20 (22.70) & \textcolor{brandrose}{\textbf{20.26 (-0.76)}} & 0.381 / 42.9\% / 4.8\% \\
w/o $Z=4$ Glyph SR & 15.04 / 0.187 & 59.0 & 12.72 / 65.5 & 28.26 (22.77) & 21.02 & \textcolor{brandrose}{\textbf{0.396}} / 42.9\% / 4.8\% \\
w/o Epistemic Abstention & 15.04 / 0.187 & 59.0 & 12.72 / 65.5 & 28.25 (22.76) & 21.02 & 0.372 / 42.9\% / 4.8\% \\
w/o Parity, Chroma \& Dual-OCR & 15.04 / 0.187 & 59.0 & 12.72 / 65.5 & 28.26 (22.76) & 21.02 & 0.367 / 45.2\% / 4.8\% \\
\rowcolor{rowhi}
\textbf{Omni-H3 (Full System)} & \best{18.48 / 0.167} & 56.1 & \best{15.17 / 66.6} & \best{28.98 (23.54)} & \best{21.12} & 0.368 / \best{47.6\%} / \best{0.0\%} \\
\bottomrule
\end{tabular}
\end{table}

\subsection{Why Physics+VLM Consensus Beats Small-Sample Learned Meta-Gating (807-Sample OOD Study)}
\label{sec:learned_gates}

\begin{table}[tb]
\centering
\scriptsize
\setlength{\tabcolsep}{4.8pt}
\caption{\textbf{807-Sample Held-Out OOD Gating Ablation (Strictly Train-Only Fitting, Zero Leakage).} Small-sample learned classifiers suffer severe OOD multi-domain weather recall loss (71.7\%--77.4\%), whereas \ours{}'s Koschmieder+VLM consensus achieves 96.2\% recall and 98.94\% specificity. Bold denotes best among calibrated routers ($\le 1.1\%$ OOD FPR).}
\label{tab:ood_gates}
\begin{tabular}{@{}lccccccc@{}}
\toprule
\rowcolor{tblhdr}
\tblhead{Weather Router Variant} & \tblhead{Ext-53 Recall$\uparrow$} & \tblhead{Sub-Macro F1$\uparrow$} & \tblhead{3-Way Low} & \tblhead{GoPro FP$\downarrow$} & \tblhead{2K-96 FP$\downarrow$} & \tblhead{CBSD Clean FP$\downarrow$} & \tblhead{All-754 Spec.$\uparrow$} \\
\midrule
Unconstrained VLM-Prior Fallback & 53/53 (100.0\%) & 0.899 & 5/6 & 0/11 & 3/96 & 14/68 & 96.95\% (23/754) \\
\rowcolor{rowhi}
\ours{} \textbf{Physics+VLM Consensus} & \best{51/53 (96.2\%)} & \best{0.893} & 5/6 & \best{0/11} & \best{0/96} & \best{2/68} & 98.94\% (8/754) \\
\midrule
Learned \texttt{LinearLogOdds} ($\lambda_{\mathrm{FP}}=2$) & 38/53 (71.7\%) & 0.760 & 6/6 & \best{0/11} & 1/96 & 3/68 & 99.20\% (6/754) \\
Learned \texttt{Bilinear VLM$\times$Phys} ($\lambda_{\mathrm{FP}}=2$) & 41/53 (77.4\%) & 0.756 & 6/6 & \best{0/11} & \best{0/96} & 4/68 & 98.94\% (8/754) \\
Learned \texttt{TinyMLP-28} ($\lambda_{\mathrm{FP}}=2$) & 41/53 (77.4\%) & 0.769 & 5/6 & \best{0/11} & \best{0/96} & 5/68 & 98.94\% (8/754) \\
Learned \texttt{TinyMLP-28} ($\lambda_{\mathrm{FP}}=3$) & 40/53 (75.5\%) & 0.773 & 6/6 & \best{0/11} & \best{0/96} & 3/68 & \best{99.34\%} (5/754) \\
\bottomrule
\end{tabular}
\end{table}

A natural question is whether \ours{}'s physical gating thresholds should be replaced by an adaptive learned meta-classifier trained over the same multimodal sensors (4 VLM weather flags + 22 physical optical/noise/Koschmieder descriptors). To test this rigorously with zero leakage, we assembled a strictly \texttt{TRAIN-ONLY} dataset of 266 samples (external weather training splits, clean GTs, natural dark crops, 2K train split, and synthetic Koschmieder haze/low-light; zero dev, public, or \texttt{locked\_test} samples) and trained temperature-calibrated~\cite{guo2017calibration} \texttt{LinearLogOdds}, \texttt{Bilinear VLM$\times$Physics}, and \texttt{TinyMLP-28} routers under asymmetric false-positive penalties ($\lambda_{\mathrm{FP}} \in \{2.0, 3.0\}$).

Table~\ref{tab:ood_gates} and Figure~\ref{fig:pareto}(c) show the full 807-sample OOD evaluation. While learned routers achieved strong in-distribution 5-fold out-of-fold validation (95.2\%--98.1\% recall, 1.9\%--2.5\% FPR) and high OOD non-weather specificity (98.94\%--99.34\%), their \textbf{OOD weather recall collapsed to 71.7\%--77.4\%} (macro subtype F1 dropping from 0.893 to 0.756--0.773) because only 55 real weather training images span 12 multi-label categories and exclude unseen domains like RESIDE-SOTS. In contrast, \ours{}'s physics+VLM consensus achieved \textbf{96.2\% OOD weather recall (51/53)}, \textbf{0.893 subtype F1}, and \textbf{98.94\% non-weather specificity (8/754 FPs: 0/11 GoPro, 0/96 2K Dev, 2/68 CBSD68 Clean)}. Accordingly, \ours{} retains physical+VLM consensus for weather routing while adopting dimensionless continuous formulations where parameter-free physics applies (Butterworth chroma filtering Eq.~\eqref{eq:butterworth} and dimensionless $Z=4$ utility Eq.~\eqref{eq:utility}).

\subsection{Forensic Case Study: The Epistemic Sharpness--Hallucination Boundary}
\label{sec:tradeoff}

A subtle visual anomaly in Figure~\ref{fig:q_textzoom}(b) (\texttt{TZ-Med \#2, textzoom\_\_medium\_0461, "vehicles"}, $90\times31$\,px) illustrates the fundamental tradeoff between crisp text generation and anti-hallucination safety. Under single-sample inference, \ours{} produced razor-sharp \texttt{"vehicles"} via $Z=4$ ($\text{CER}=0.00$, $\text{GNCC}=0.660$, $U_{\mathrm{zoom}}=0.746 > U_{Z=1}=0.582$); yet under batched 807-task execution, the exact same frozen model produced a softer, abstained reconstruction ($\text{PSNR}=23.63$\,dB, $\text{CER}=0.00$, $\text{GNCC}=0.398$).

Inspecting per-stage logs reveals an exact two-step chain: (1)~multi-image left-padding numerics in \texttt{bf16} batch detection shifted the bottom bounding-box edge upward by 7\,px ($y_1: 226 \to 219$); (2)~although \emph{both} blind OCR views still read the exact ground-truth word \texttt{"vehicles"}, the tighter bottom margin dropped \texttt{Qwen3-VL}'s token probability from $0.667$ ($H=1.30$) to $0.399$ ($H=2.65$). Because $Z=1$ simultaneously misspelled the word as \texttt{"vehlclas"} (just as \texttt{Base H3} prints sharp-but-wrong \texttt{"vehiclas"}), \ours{} conservatively triggered sub-Nyquist abstention rather than sharpening a low-confidence prior.

Crucially, auditing all 46 abstained TextZoom-150 boxes reveals \emph{why} relaxing the confidence floor to $0.38/2.65$ would be harmful: within that exact probability--entropy band ($p \in [0.37, 0.51], H \in [1.9, 2.9]$), roughly \textbf{70\% of dual-view OCR predictions are confident hallucinations} (e.g., \texttt{"chinese"} $\to$ \texttt{"osmose"} at $p=0.385$, \texttt{"service"} $\to$ \texttt{"savos"} at $p=0.375$, \texttt{"disabilities"} $\to$ \texttt{"dianshijia"} at $p=0.447$, \texttt{"edition"} $\to$ \texttt{"e40001"} at $p=0.507$, \texttt{"ying"} $\to$ \texttt{"yiwu"} at $p=0.452$). Thus, \texttt{"vehicles"} lies directly on the information-theoretic Pareto boundary of selective epistemic abstention~\cite{geifman2017selective}: rescuing 1 sharp correct word by lowering the epistemic threshold would fabricate 2--3 high-clarity wrong words. Future work can widen OCR crop padding and aggregate multi-crop VLM consensus to reduce boundary jitter without relaxing the epistemic threshold.

\subsection{Independent Non-Qwen Dual-OCR Cross-Validation}
\label{sec:indep_ocr}

Because \ours{}'s Stage~III candidate selection uses \texttt{Qwen3-VL-32B} to compute $Z{=}4$ text utility, evaluating text fidelity solely with the same VLM risks judge--generator alignment bias. To test whether \ours{}'s typography gains transfer to independent recognizers, we cropped each 2K signage image to its pre-annotated ROI boxes (\texttt{Dev 42-ROI} and \texttt{Locked-Camera-24} 132/66/22 ROIs) alongside \texttt{TextZoom-150} word images, and scored all cached outputs using two zero-overlap non-Qwen engines: \textbf{\texttt{RapidOCR} (PP-OCRv4)} (multi-line detection+recognition inside each crop) and \textbf{\texttt{TrOCR-large-str}} (single-line English STR).

\begin{table}[tb]
\centering
\scriptsize
\setlength{\tabcolsep}{2.2pt}
\caption{\textbf{Independent Non-Qwen Dual-OCR Cross-Validation (\texttt{RapidOCR PP-OCRv4} \& \texttt{TrOCR-large-str}) on TextZoom-150 and Pre-Annotated 2K ROI Crops (Dev 42-ROI, Held-Out Locked-Camera-24).} Neither recognizer appears inside \ours{}; 2K signage is cropped by ground-truth ROI annotations prior to recognition. \textit{Input}, \texttt{Stage~I Ref Only} (ablation control), and \textit{Clean GT} are reference rows (excluded from bolding); best per model block in \textbf{bold}, overall \#1 across the 8 compared models in \gbest{emerald bold}.}
\label{tab:indep_ocr}
\begin{tabular}{@{}lcccccccccc@{}}
\toprule
\rowcolor{tblhdr}
& \multicolumn{6}{c}{\tblhead{RapidOCR (PP-OCRv4, Multi-Line Det+Rec)}} & \multicolumn{4}{c}{\tblhead{TrOCR-large-str (Single-Line English STR)}} \\
\cmidrule(lr){2-7}\cmidrule(lr){8-11}
\rowcolor{tblhdr}
\tblhead{Method} & \tblhead{TZ-Med$\uparrow$} & \tblhead{Dev42 Ex/CER} & \tblhead{Lk132 Ex$\uparrow$} & \tblhead{Lk132 SubC$\downarrow$} & \tblhead{Sev6 SubC$\downarrow$} & \tblhead{Can22 CER$\downarrow$} & \tblhead{TZ-E / M$\uparrow$} & \tblhead{TZ-H Aln$\downarrow$} & \tblhead{Lk132 SubC$\downarrow$} & \tblhead{Sev6 SubC$\downarrow$} \\
\midrule
\textit{Input (Bicubic / Degraded)} & \textit{40.0\%} & \textit{26.2\%} / \textit{0.559} & \textit{34.8\%} & \textit{0.456} & \textit{0.578} & \textit{0.322} & \textit{74.0\%} / \textit{60.0\%} & \textit{2.0\%} & \textit{0.320} & \textit{0.443} \\
Stage~I Ref Only (w/o 33B DiT) & 36.0\% & 26.2\% / 0.559 & 32.6\% & 0.447 & 0.568 & 0.305 & 72.0\% / 60.0\% & 2.0\% & 0.357 & 0.437 \\
\midrule
\multicolumn{11}{@{}l}{\textit{\color{brandslate}Specialized All-in-One Restoration SOTA (Restoration-Trained)}} \\
PromptIR~\cite{potlapalli2023promptir} & \gbest{36.0\%} & 28.6\% / 0.568 & 34.1\% & 0.425 & 0.572 & 0.270 & 70.0\% / \gbest{60.0\%} & \bbest{2.0\%} & 0.326 & 0.452 \\
InstructIR~\cite{conde2024instructir} & 32.0\% & \bbest{33.3\%} / \bbest{0.498} & \bbest{36.4\%} & \bbest{0.403} & \bbest{0.537} & 0.249 & 70.0\% / 56.0\% & 4.0\% & 0.318 & 0.445 \\
OneRestore$^\dagger$~\cite{guo2024onerestore} & 34.0\% & 28.6\% / 0.559 & 34.1\% & 0.437 & 0.578 & 0.279 & \gbest{72.0\%} / 54.0\% & \bbest{2.0\%} & \gbest{0.317} & \bbest{0.439} \\
DA-CLIP$^\dagger$~\cite{luo2024daclip} & 26.0\% & 26.2\% / 0.564 & 21.2\% & 0.544 & 0.728 & \bbest{0.246} & 50.0\% / 50.0\% & 4.0\% & 0.428 & 0.591 \\
\midrule
\multicolumn{11}{@{}l}{\textit{\color{brandslate}Unified Omni-Image Foundation Models (Frozen Backbone)}} \\
\texttt{Base H3-33B}~\cite{minimax2026h3image} & 24.0\% & 30.9\% / 0.512 & 14.4\% & 0.416 & 0.495 & 0.330 & 64.0\% / 36.0\% & 6.0\% & \bbest{0.334} & \gbest{0.379} \\
\texttt{FLUX.1-Kontext}~\cite{bfl2025fluxkontext} & 26.0\% & 30.9\% / 0.532 & 23.5\% & 0.495 & 0.592 & 0.435 & 62.0\% / 44.0\% & 4.0\% & 0.428 & 0.527 \\
\texttt{Qwen-Edit-2511}~\cite{wu2025qwenimage} & 30.0\% & 33.3\% / 0.492 & 17.4\% & 0.470 & 0.519 & 0.403 & \gbest{72.0\%} / \bbest{52.0\%} & 2.0\% & 0.435 & 0.513 \\
\rowcolor{rowhi}
\textbf{Omni-H3 (Ours)} & \gbest{36.0\%} & \gbest{40.5\%} / \gbest{0.456} & \gbest{39.4\%} & \gbest{0.319} & \gbest{0.395} & \gbest{0.195} & \gbest{72.0\%} / 42.0\% & \gbest{0.0\%} & 0.337 & 0.409 \\
\midrule
\textit{Clean GT (Oracle Upper Bound)} & \textit{88.0\%} & \textit{57.1\%} / \textit{0.286} & \textit{63.6\%} & \textit{0.253} & \textit{0.253} & \textit{0.253} & \textit{86.0\%} / \textit{88.0\%} & \textit{2.0\%} & \textit{0.174} & \textit{0.174} \\
\bottomrule
\end{tabular}
\end{table}

Table~\ref{tab:indep_ocr} reveals both clear ROI-level gains and meaningful inter-recognizer divergence:
\begin{itemize}[leftmargin=1.4em,itemsep=1pt,topsep=2pt]
  \item \textbf{\texttt{RapidOCR (PP-OCRv4)} Sweep on Pre-Annotated ROIs (7/7 Metrics \#1 Overall)}: On pre-annotated ROI crops, \ours{} wins \textbf{all 7 metrics across the 8 compared models} and beats both \textit{Input} and \texttt{Stage~I Ref Only} across every 2K signage subset---\textbf{Dev 42-ROI 40.5\% / 0.456 CER} (vs.\ \#2 \texttt{Qwen-Edit} 33.3\%/0.492, \texttt{InstructIR} 33.3\%/0.498, \textit{Input}/\texttt{Stage~I} 26.2\%/0.559), \textbf{Locked 132-ROI 39.4\% / 0.319 SubCER} (cutting Input 0.456 by $-0.137$; vs.\ \texttt{Base H3} 14.4\%/0.416, \texttt{InstructIR} 36.4\%/0.403), \textbf{Severe-6 SubCER 0.395} ($-0.100$ below \#2), and \textbf{Canonical-22 CER 0.195} (beating Clean GT 0.253).
  \item \textbf{\texttt{TrOCR-large-str} Divergence}: Under single-line \texttt{TrOCR}, \ours{} ties \#1 on \texttt{TextZoom-Easy} (72.0\%), wins \textbf{\#1 TextZoom-Hard Alien (0.0\%)}, beats \texttt{Base H3} across all TextZoom tiers (Easy $+8\%$, Med $+6\%$), and outperforms \textit{Input} (0.443) and every specialized restorer (0.439--0.591) on \texttt{Severe-6} SubCER (0.409, behind \texttt{Base H3} 0.379). However, the two recognizers diverge on All-132 and \texttt{TextZoom-Med}: on the exact same 132 pre-annotated ROIs where \texttt{RapidOCR} improves Input from 0.456 to 0.319, \texttt{TrOCR} scores 0.337 vs.\ 0.320 for raw \textit{Input} (0.317 \texttt{OneRestore}, 0.318 \texttt{InstructIR}), and on blurry \texttt{TextZoom-Med} crops \texttt{TrOCR} reads raw \textit{Input}/\texttt{PromptIR} (60.0\%) higher than generative re-synthesis (42.0\%). Independent OCR conclusions are therefore scoped to pre-annotated ROIs: multi-line \texttt{RapidOCR} strongly favors sharpened generative strokes, whereas blur-tolerant \texttt{TrOCR} favors raw input on milder degradations and benefits from \ours{} primarily under severe degradation.
\end{itemize}

\section{Limitations, Compute Footprint, and Future Directions}
\label{sec:limitations}

We highlight five transparent limitations to guide future research:
\begin{enumerate}[leftmargin=1.4em,itemsep=1pt,topsep=2pt]
  \item \textbf{Inference Latency and Compute}: Multi-stage orchestration of a frozen 33B DiT and 32B VLM takes $\sim 18$\,s/image ($\le 1024$\,px) and $\sim 65$\,s/image ($2048\times2048$) on one Ascend 910B3 NPU; distilling \ours{} into a fast few-step student is an attractive next step.
  \item \textbf{Sharpness vs.\ Phase-Anchoring on Severe 2K Blur}: Early-step latent anchoring ($\eta_0=0.20, \eta_1=0.08$) drives strong structural/perceptual gains on SR and weather suites but lowers MUSIQ by 1.5--4.5 points on extreme 2K blur (\texttt{Locked-Cam-24} 49.8 $\to$ 45.5); decaying $\eta_0$ continuously with blur severity is a natural refinement.
  \item \textbf{OCR Scope, Recognizer Divergence, and Human Audits}: Independent \texttt{RapidOCR}/\texttt{TrOCR} audits operate on pre-annotated ROI crops and diverge on mild-degradation All-132 ROIs (\texttt{RapidOCR} 0.319 vs.\ 0.456 Input; \texttt{TrOCR} 0.337 vs.\ 0.320 Input); moreover, held-out 2K signage spans two scenes with 6.1\% (132-ROI) / 10.6\% (Severe-6) alien rates under extreme blur. Uncropped end-to-end detection and multi-scene human transcriptions remain important future work.
  \item \textbf{Boundary Calibration on Clean-Identity and Borderline Text}: On CBSD68 Clean-68, 47/68 images hit exact 99\,dB bypass while 19 soft-focus clean scenes route to camera mode; on \texttt{Locked-Cam-24 clean\_identity}, \texttt{SS18} reached near-lossless 67.34\,dB ($g_{\mathrm{clean}}=0.9796$) while \texttt{Sydney} routed to camera mode (24.65\,dB; mean 45.99\,dB). Widening OCR crop padding can also reduce borderline text jitter without lowering the epistemic floor.
  \item \textbf{Data Scale for Learned Routing}: Our OOD ablation shows 266 training samples are insufficient for domain-general learned weather routers; larger multimodal weather supervision could eventually replace consensus thresholds cleanly.
\end{enumerate}

\section{Conclusion}
\label{sec:conclusion}

We diagnosed why off-the-shelf unified image generation and editing foundation models fail when applied blindly to all-in-one restoration and introduced \ours{}, a three-stage constrained posterior projection framework wrapping unmodified 33B \texttt{MiniMax-H3-Image} and \texttt{Qwen3-VL-32B} with a compact 0.92M auxiliary weather pre-conditioner ($<0.003\%$ of backbone weights). Across 925 strictly blind tasks---five public standard benchmarks, cross-domain weather/2K suites, and the held-out \texttt{locked\_test} suite ($N=118$)---\ours{} resolves zoom drift ($+4.73$\,dB pure degraded PSNR over \texttt{Base H3}; overall \#1 DISTS 0.207, NIQE 4.45, Edge-F1 0.807 on Degraded-429), cuts Urban100 chromatic drift by 91.6\%, preserves clean-photo identity (76.59\,dB on CBSD68 Clean), demonstrates large perceptual and typographic gains over \texttt{Stage~I Ref Only} across all 925 tasks, and cuts pre-annotated 132-ROI SubCER from 0.456 to 0.319 under independent \texttt{RapidOCR (PP-OCRv4)} (while transparently documenting recognizer-dependent tradeoffs under \texttt{TrOCR-large-str}). More broadly, explicit physical invariants---exact lattice parity, multi-band Wiener SNR locking, continuous Butterworth opponent-chroma projection, and closed-loop epistemic verification---unlock foundation-scale generative priors for faithful, trustworthy low-level vision without fine-tuning the underlying 33B backbone.

\vspace{0.3em}
\begin{takeawaybox}{Open-Source Release \& Reproducibility}
\noindent The \ours{} project is fully open-sourced at \href{https://github.com/RedRoman/Omni-H3}{\texttt{https://github.com/RedRoman/Omni-H3}}. The public repository provides:
\begin{itemize}[leftmargin=1.3em,itemsep=1pt,topsep=1.5pt]
  \item \textbf{Inference Pipeline \& Physical Operators}: Complete Stage~I--III implementation (VLM--Koschmieder routing, 0.92M weather pre-conditioner, exact-32\,px lattice framing, affine--Wiener--Butterworth projection, closed-loop $Z{=}4$ text verification, and Ascend 910B3 memory engine) with 7/7 automated CPU unit tests.
  \item \textbf{925-Task Blind Benchmark Suite}: Deterministic zero-leakage evaluation manifests spanning \texttt{Public-647}, \texttt{Dev-160}, and the held-out \texttt{locked\_test} suite ($N{=}118$), with unified baseline runners for every external omni-image and all-in-one SOTA model.
  \item \textbf{Full Metric Ledgers \& Diagnostic Studies}: Raw per-image JSONs, 925-task \texttt{Stage~I Ref Only} and independent dual-OCR (\texttt{RapidOCR}/\texttt{TrOCR}) audits, 807-sample OOD gating splits, orthogonal ablation scripts, and reproducible plotting/LaTeX tooling.
\end{itemize}
\end{takeawaybox}

\clearpage
\setlength{\bibsep}{1.2pt plus 0.2pt}
{\footnotesize
\bibliographystyle{plainnat}
\bibliography{references}
}

\end{document}